\documentclass[10pt,twocolumn,letterpaper]{article}

\usepackage[pagenumbers]{cvpr} % To force page numbers, e.g. for an arXiv version

\usepackage{graphicx}
\usepackage{amsmath}
\usepackage{amssymb}
\usepackage{booktabs}

\usepackage{multirow}
\usepackage{rotating}
\usepackage{siunitx}
\usepackage{subcaption}
\usepackage{textcomp}

\usepackage[pagebackref,breaklinks,colorlinks]{hyperref}

\usepackage[capitalize]{cleveref}
\crefname{section}{Sec.}{Secs.}
\Crefname{section}{Section}{Sections}
\Crefname{table}{Table}{Tables}
\crefname{table}{Tab.}{Tabs.}
\crefname{figure}{Fig.}{Figs.}
\Crefname{figure}{Figure}{Figures}

\newcommand\blfootnote[1]{%
	\begingroup
	\renewcommand\thefootnote{}\footnote{#1}%
	\addtocounter{footnote}{-1}%
	\endgroup
}

\usepackage[final]{microtype}

\def\cvprPaperID{13} % *** Enter the CVPR Paper ID here
\def\confName{CVPR Workshop on 3D Vision and Robotics}
\def\confYear{2023}

\begin{document}

%%%%%%%%% TITLE - PLEASE UPDATE
\title{Reviewing 3D Object Detectors in the Context of High-Resolution 3+1D Radar}

\author{Patrick Palmer\textsuperscript{1}*, Martin Krueger\textsuperscript{1}*, Richard Altendorfer\textsuperscript{2}, Ganesh Adam\textsuperscript{2}, Torsten Bertram\textsuperscript{1}\\
\textsuperscript{1} TU Dortmund University, Germany \hspace{0.5cm} \textsuperscript{2} ZF Group\\
{\tt\small \{patrick.palmer, martin2.krueger, torsten.bertram\}@tu-dortmund.de}\\
{\tt\small \{richard.altendorfer, ganesh.adam\}@zf.com}
% For a paper whose authors are all at the same institution,
% omit the following lines up until the closing ``}''.
% Additional authors and addresses can be added with ``\and'',
% just like the second author.
% To save space, use either the email address or home page, not both
}
\maketitle

%%%%%%%%% ABSTRACT
\begin{abstract}
   Recent developments and the beginning market introduction of high-resolution imaging 4D (3+1D) radar sensors have initialized deep learning-based radar perception research. We investigate deep learning-based models operating on radar point clouds for 3D object detection. 3D object detection on lidar point cloud data is a mature area of 3D vision. Many different architectures have been proposed, each with strengths and weaknesses. Due to similarities between 3D lidar point clouds and 3+1D radar point clouds, those existing 3D object detectors are a natural basis to start deep learning-based 3D object detection on radar data. Thus, the first step is to analyze the detection performance of the existing models on the new data modality and evaluate them in depth. In order to apply existing 3D point cloud object detectors developed for lidar point clouds to the radar domain, they need to be adapted first. While some detectors, such as PointPillars, have already been adapted to be applicable to radar data, we have adapted others, \eg, Voxel R-CNN, SECOND, PointRCNN, and PV-RCNN.
   To this end, we conduct a cross-model validation (evaluating a set of models on one particular data set) as well as a cross-data set validation (evaluating all models in the model set on several data sets). The high-resolution radar data used are the View-of-Delft and Astyx data sets. Finally, we evaluate several adaptations of the models and their training procedures. We also discuss major factors influencing the detection performance on radar data and propose possible solutions indicating potential future research avenues.
\end{abstract}

% https://github.com/BritishMachineVisionAssociation/BMVCTemplate/issues/11
\blfootnote{$^\ast$ Equal Contribution.}

%%%%%%%%% BODY TEXT
\vspace{-0.75cm}
\section{Introduction} \label{sec:introduction}
\subsection{Perception} \label{subsec:perception}
The three most common exteroceptive sensors currently used for automated driving tasks are camera, lidar, and radar. Camera sensors use sequential images (video) to capture the scene. Cameras have the advantage that they are comparatively cheap and widely used in different domains. Another benefit is that the camera signals are easily interpretable by humans, allowing for an easy examination of detection results. One negative aspect of camera sensors is that they do not allow for a precise measurement of distances and velocities. Lidar sensors use laser beams and measure the time-of-flight of reflected beams from detected objects. The advantages of this sensor type are the very accurate range measurement and its possibility to get a comparatively dense representation of the scene as a point cloud. Adversely for lidar is its high costs, which prevented its use in mass-production vehicles until recently. Some of these problems can be solved with radar sensors. Compared to camera and lidar sensors, radar has unique benefits. Camera and lidar sensors provide a high angular resolution but suffer in view range, whereas radar exceeds them and can therefore supplement the other sensors. Radar sensors provide a direct measurement of the Doppler velocity. This can be used to separate moving objects from one another and to distinguish static objects. The large wavelength of the radar is advantageous in adverse weather conditions like snow, rain, fog, or poor lighting conditions, where the other sensors could suffer. Radar sensors are also cost-efficient. One problem of current series production radar sensors is the sparsity of the measurements. This issue is alleviated by current advances in high-resolution 3+1D radar technology, which enable an increased field-of-view and a higher elevation resolution.

Different perception tasks are typically investigated for camera and point cloud data (lidar and radar). However, we only focus on 3D bounding box detection in this paper.

\subsection{Radar} \label{subsec:radar}
\begin{figure*}[htbp]
	\centering
	\resizebox{16cm}{!}{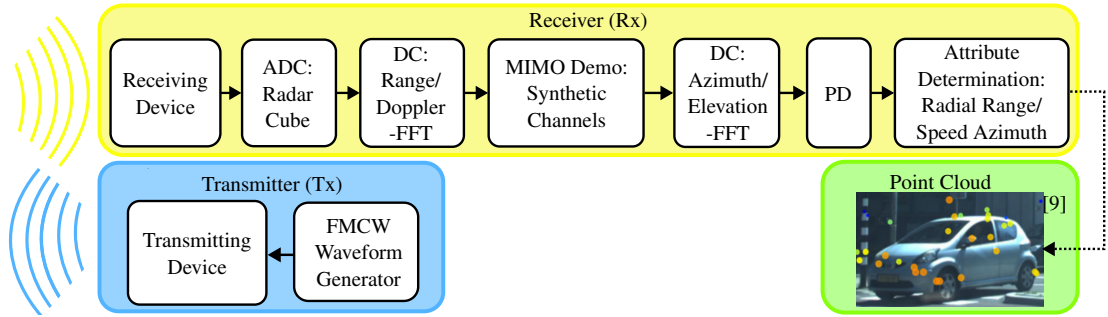}
	\caption{Overview of radar sensor data processing pipeline. See~\cite{Radar_AD_19,Automotive_Radar_20,RadarProcessing_17} for a more detailed description. ADC: Analog-to-Digital Converter, FFT: Fast Fourier Transformation, DC: Data Conversion, Demo: Demodulation, PD: Peak Detection}
	\label{fig:radar_data_processing}
	\vspace{-0.375cm}
\end{figure*}

\textbf{Data Processing} \hspace{0.25cm}
Radar sensors use electro-magnetic waves in the radio waves spectrum of 24GHz as well as 77-81GHz. In order to determine the distance to objects, the radio frequency must be varied; the most common method is the periodic continuous frequency variation (Frequency Modulated Continuous Wave: FMCW). Angular resolution depends upon the number of transmitting and receiving antennae whose combinations form virtual channels for provision of angular information. \cref{fig:radar_data_processing} shows the high-level building blocks of a typical 3+1D (radial range, radial velocity, azimuth angle, elevation angle) automotive radar sensor.

\textbf{Radar Data Representation Formats} \hspace{0.25cm} There are two main representations of radar measurements that are used for object detection, the Range-Azimuth-(Elevation-)Doppler (RA(E)D) spectrum, which is derived from the raw radar time signal using a Fast Fourier Transformation (FFT), and the point cloud representation. The point cloud can be derived from the RA(E)D spectra using, for example, a Constant False Alarm Rate (CFAR) detector~\cite{CFAR_Reference}. Both representations have unique benefits. The RA(E)D spectra contain more information than the point cloud but require more computational resources and larger data bandwidths. Point clouds, on the other hand, have the advantage of being more computationally efficient and are widely used in lidar point cloud detection. Since we want to directly adapt those models to radar point cloud data, we neglect models working on RA(E)D spectrum data for the rest of our paper.

\subsection{Contribution} \label{subsec:contribution}
As a conclusion from the previous thoughts, the following research question is raised: \textit{How accurate are existing 3D object detectors compared on 3+1D high-resolution radar point cloud data?} Therefore, our main contributions are:
\begin{itemize}
	\item adapt point-, voxel- and point-voxel-based 3D object detectors and their respective training configurations to radar data (including but also extending previous adaptations of the Voxel Feature Encoder (VFE)~\cite{ViewOfDelftDataset_22}),
	\item training ten 3D object detectors on the View-of-Delft (VoD) data set~\cite{ViewOfDelftDataset_22} and evaluating them deeply, and
	\item fine-tune the trained models on the Astyx data set~\cite{AstyxDataset_19} and conduct a detailed evaluation.
\end{itemize}

\section{State-of-the-Art for 3D Point Cloud Detection Approaches} \label{sec:state-of-the-art}
According to the surveys~\cite{LidarSurvey_21, 3DDetectionSurvey_22} on lidar point cloud-based object detection, the following architectures can be distinguished: point-based, voxel-based, pillar-based, and dual representation-based (point-voxel and point-pillar-based), classified by the data processing format.

On the other hand, \cite{RadarDetectionComparison_21} compares different detection approaches on radar data. The presented and evaluated models are mostly hand-crafted and contain much feature-engineering such as the definition of the three input channels of the grid map for a 2D CNN-based detector. While PointPillars~\cite{Pointpillar} seems to be the most popular model of the family of pillar-based architectures in~\cite{LidarSurvey_21}, the other models from~\cite{RadarDetectionComparison_21} cannot be easily assigned to typical end-to-end trainable model families from~\cite{LidarSurvey_21, 3DDetectionSurvey_22}. Two of the evaluated models contain PointNet++~\cite{Pointnet_plus} and the YOLOv3-based~\cite{Redmon_Yolov3} architecture similar to PIXOR~\cite{PIXOR_18}.

Recent approaches like CenterFusion~\cite{CenterFusion_21} and PointPillars-Radar~\cite{ViewOfDelftDataset_22} demonstrate that 3D point cloud detectors initially developed for lidar perception can be adapted to high-resolution imaging radar point cloud data.

We focus our investigation on established models initially developed for lidar. However, acknowledging the effort and valuable insights of the investigations from~\cite{RadarDetectionComparison_21}, we would like to extend their comparison of different detectors on radar data. We also want to relate our results to the general findings for lidar~\cite{LidarSurvey_21} and radar object detection~\cite{RadarDetectionComparison_21}.

Next, the three classes of point cloud detection models defined by~\cite{LidarSurvey_21} and later used in our experimental evaluation are briefly introduced. Five of the models are two-stage, and five are one-stage detectors. Additionally, general patterns characterizing their lidar perception results are provided. This information should be considered later when evaluating the performance on the radar data.

\subsection{Point-based} \label{subsec:point-based}
Point-based methods usually follow the classic pipeline of alternately down-sampling the original point cloud, encoding in the backbone network, and finally, applying a detection head. For feature encoding or learning in the backbone network, PointNet~\cite{PointNet} or PointNet++~\cite{Pointnet_plus} are often used, applying a cascade of feature aggregation with Multilayer Perceptrons (MLPs) and max-pooling layers to learn local structures. PointNet modules are often used to build an encoder-decoder structure within specific architectures.

\textbf{Point-RCNN} \hspace{0.25cm} PointRCNN~\cite{PointRCNN} is a two-stage object detector. The first stage of classic point-based models is extended by foreground-background segmentation. This segmentation information is then concatenated to the encoded point features for generating 3D Region-of-Interests (RoIs) that are cleaned up by Non-Maximum Suppression (NMS). In the second stage, those RoIs are extended, and another round of feature encoding is done. Finally, the detection head outputs a confidence score and a refined bounding box.

\textbf{3DSSD} \hspace{0.25cm} 3DSSD~\cite{3DSSD_20} is a point-based one-stage detector. The model successfully removes the feature propagation layers in the backbone network required by other point-based detectors for upsampling features. A fusion sampling strategy compensates for this by combining standard distance-based furthest-point-sampling (FPS) and feature-FPS. Furthermore, the proposed candidate generation layer and center-ness assignment strategy enable the removal of the refinement stage, which reduces the inference time.

\textbf{Evaluation for Lidar (and Radar) Point Clouds} \hspace{0.25cm} According to~\cite{LidarSurvey_21}, point-based detectors are generally assessed as \textit{overall satisfactory}, but their real-time capabilities are questioned due to their two-stage structure. 3DSSD \cite{3DSSD_20} as one-stage model, has a runtime of just \SI{38}{\milli\second} while reaching detection accuracies comparable to the best-performing point and voxel-based two-stage detectors. The results of the two models containing PointNet and PointNet++ from the RadarScenes data set~\cite{RadarScenesDataset_21} in~\cite{RadarDetectionComparison_21} are ambiguous since one of the models performed second best and the other one second worst. However, a detailed evaluation revealed that at least for the second-worst model, the DBSCAN-based cluster stage in this non-end-to-end trainable model accounts for the weak performance, limiting the significance of these results for evaluating PointNet and PointNet-based models.

\subsection{Voxel-based} \label{subsec:voxel-based}
Voxel-based approaches first transform the continuous point cloud into a 3D cube of equally sized discrete voxels. Afterward, the points within a voxel are encoded by a VFE. Then, the voxelized data gets processed by a 3D (sparse) convolutional backbone network before the detection header finally derives 3D detections. Sparse convolutions~\cite{Sparse_Convolutions_17} only apply their computations to the parts of the input data that are non-empty. Due to the nature of 3D lidar data, most of the derived voxels are empty, leading to an enormous computational and storage overhead if not implemented intelligently, i.e., using sparse convolutions.

\textbf{SECOND} \hspace{0.25cm} SECOND (Sparsely Embedded CONvolutional Detection)~\cite{Second} is a one-stage detection model that utilizes sparse 3D convolutional operations and introduces additional improvements during training, such as sine-error loss for yaw angle regression. Its 2D convolutional detection head consumes the output of the sparse 3D convolution backbone and then derives its object detections from an anchor-free Region Proposal Network (RPN).

\textbf{Part A\textsuperscript{2}} \hspace{0.25cm} Part A\textsuperscript{2} (part-aware and part-aggregation net)~\cite{PartA2_21} is a two-stage detector that has an anchor-free and an anchor-based configuration, both sharing the same remaining architecture. The first stage that generates detection proposals is implemented as an encoder-decoder structure using the 3D CNN UNet~\cite{UNet_15} approach, followed by a sparse 3D convolutional backbone network. In the second stage, the detection proposals are refined considering spatial relations using RoI-aware pooling and a sparse 3D CNN.

\textbf{Voxel R-CNN} \hspace{0.25cm} Voxel R-CNN~\cite{Voxel_R-CNN_21} is a two-stage detector whose first stage uses standard sparse 3D and 2D convolutional backbone networks to generate anchor-based 3D bounding box proposals. Next, a fixed-size voxel grid around each object proposal is selected, represented by its center point. Then the newly proposed voxel query operation is applied to utilize the data's structure to gain efficiency. Finally, an adaptation of PointNet~\cite{PointNet} aggregates information from the query to feed it into the final  fully connected (FC) layer to generate object detections.

\textbf{Evaluation for Lidar Point Clouds} \hspace{0.25cm} Referring to~\cite{LidarSurvey_21}, voxel-based models reach \textit{state-of-the-art or top} detection performances. Inference time depends upon two criteria. First, the larger the used 3D convolutional backbone network is, the more accurate the detections become but at the cost of inference time. Contrary, focusing more on the 2D convolutional backbone instead of the 3D modules leads to faster inference but at the cost of detection accuracy, at least in 3D evaluations. The second main influence comes with respect to using one and two-stage detectors. One-stage detectors are historically faster. Recently, with~\cite{Voxel_R-CNN_21}, a two-stage detector could reach almost an inference time of one-stage detectors with the additional benefit of enhanced accuracy. 

\subsection{Pillar-based} \label{subsec:pillar-based}
Pillars are a special form of voxels, spanned over the entire height along the $z$ dimension in a Cartesian coordinate system. Therefore, points are not sorted in separate cells according to their vertical position. Instead, all points inside a pillar are encoded, usually by a PointNet-like~\cite{PointNet} architecture. The encoded features are then interpreted as channels in a grid-map representation processed by a 2D CNN architecture to generate detections.

\textbf{PointPillars} \hspace{0.25cm} PointPillars~\cite{Pointpillar} is a one-stage detector. The VFE is based on the respective module from~\cite{VoxelNet} and takes nine point attributes as input. The 2D CNN module is followed by an anchor-free RPN that directly outputs a 3D bounding box and a confidence score for each classification.

\textbf{CenterPoint} \hspace{0.25cm} CenterPoint~\cite{CenterPoint_21} is a two-stage point cloud detection approach that operates on pillar or voxel data representation. The pillar-based implementation uses PointPillars to construct a bird's-eye-view (BEV) intermediate representation fed into an anchor-free region-proposal head for 3D bounding box regression. Next, the center points of the proposals and four additional points indicating the center of each face of the BEV object proposal are concatenated and then fed to the final terminal head that generates the final detections. We decided to use the pillar-based implementation since~\cite{ViewOfDelftDataset_22} reached good results on radar data with their PointPillars-based model. We call the radar model CenterPoint-R using PointPoillars-Radar (PointPoillars-R) as a base model and the lidar model CenterPoint-L utilizing the original PointPillar model~\cite{Pointpillar}.

\textbf{Evaluation for Lidar (and Radar) Point Clouds} \hspace{0.25cm} Since pillar-based modules avoid using 3D CNN layers and instead only use 2D CNN structures, they are computationally more efficient. According to~\cite{LidarSurvey_21}, this comes at the cost of an inferior detection accuracy on KITTI lidar data, at least for the difficult category. They also emphasize that pillar-based models perform worse than voxel-based ones on more complex lidar data sets such as Waymo~\cite{WaymoDataset_20}. \cite{LidarSurvey_21} attribute this degradation to the simplistic VFE and the fact that a 2D CNN backbone cannot capture the rich structure of a 3D point cloud which requires 3D CNN layers. Pillar-based models can reach satisfactory results on the easy and moderate examples on KITTI lidar data, according to~\cite{LidarSurvey_21}.

\subsection{Dual Representation-based} \label{subsec:dual_representation-based}
Dual representation-based approaches try to combine the benefits from their respective model families. While voxel-based detectors are computationally efficient, point-based models can leverage detailed structural information. In one-stage models, voxel-based and point-based architectures are processed in parallel and share information in voxel-to-point or point-to-voxel modules ~\cite{LidarSurvey_21}. Two-stage models such as PV-RCNN~\cite{PV_RCNN} generate object proposals in the first stage by a voxel-based architecture and refine their detections in the second stage based on the proposed keypoints.

\textbf{PV-RCNN} \hspace{0.25cm} In the first stage, PV-RCNN~\cite{PV_RCNN} uses the SECOND~\cite{Second} model to generate 3D bounding boxes and assign keypoints to them. In the second stage, the previously determined voxel features are concatenated with two newly generated feature vectors provided by a PointNet-based branch and a 2D backbone network based on a BEV representation of the scene. Additionally, the keypoints are weighted according to a foreground-background segmentation module to boost performance further.

\textbf{Evaluation for Lidar Point Clouds} \hspace{0.25cm} Two-stage detectors reach state-of-the-art detection results on the KITTI lidar data, according to~\cite{LidarSurvey_21}. Their detection performance improves slightly compared to voxel-based models, while the inference time increases only marginally.

The presented model overview is non-exhaustive. We instead focused our model review on the most popular and promising model architectures.

\section{Experimental Evaluation} \label{sec:experimental_evaluation}
\subsection{Experimental Setup} \label{subsec:experimental_setup}
Before presenting and discussing the results of our investigation, the experimental procedure is introduced. The considered models are the ones described in \cref{subsec:point-based,subsec:voxel-based,subsec:pillar-based,subsec:dual_representation-based}.

\textbf{Implementation Details} \hspace{0.25cm} We used existing implementations of the evaluated models provided by the OpenPCDet framework\footnotemark[1]\footnotetext[1]{\url{https://github.com/open-mmlab/OpenPCDet}}. We had to adapt voxel sizes to feed the radar data into the different detectors. Like~\cite{ViewOfDelftDataset_22}, we used KITTI~\cite{KITTIDataset_13} models of OpenPCDet as initial implementation. Adopting the relevant point cloud area in the $x-y$ plane from~\cite{ViewOfDelftDataset_22} requires shrinking the voxel size to $0.036 \times 0.032 \times 0.125$\SI{}{\meter} if the 3D object detection models should stay unchanged. Additionally, we also implemented a model operating on larger voxels ($0.135 \times 0.120 \times 0.625$\SI{}{\meter}) for the radar data. When not explicitly mentioned, the configuration with smaller voxel sizes was used in our experiments. For the PointPillars-R model\footnotemark[2]\footnotetext[2]{\url{https://github.com/tudelft-iv/view-of-delft-dataset/blob/main/PP-Radar.md}} we followed the guidelines given in~\cite{ViewOfDelftDataset_22}, integrating the model and supporting code into OpenPCDet. For clearer discrimination, we explicitly call the standard PointPillars~\cite{Pointpillar} PointPillars-Lidar (PointPillars-L) to emphasize the lidar configuration. The necessary adaptations for the remaining models to be compatible with radar measurements, including additional features such as relative radial velocity or Radar Cross Section (RCS), have been done according to the example of~\cite{ViewOfDelftDataset_22}. In addition, training procedure modifications were necessary to achieve better results. OpenPCDet does not provide learning rate (lr) schedulers that reduce the lr with respect to the learning progress using the loss or a validation error metric. Nevertheless, we used its implementation of the one-cycle lr scheduler~\cite{One-Cycle_Schedules_18}. When we modified the standard configuration from OpenPCDet to reach the maximal lr earlier and extended the decaying phase later, we got better results. Therefore, as the theory supports, a large lr regularizes the training in the high-dimensional optimization landscape~\cite{One-Cycle_Schedules_18}. Later, when an appropriate area in that landscape has been found, decaying the lr helps to converge to a local minimum smoothly.

\textbf{Data} \hspace{0.25cm} We evaluate the mentioned models on two data sets: the View-of-Delft (VoD)~\cite{ViewOfDelftDataset_22} and the Astyx~\cite{AstyxDataset_19} data set. \textbf{VoD} is our main data set since it is significantly larger. We use it to train all the models from scratch, using the provided data set split. Since the separate test set of VoD does not contain annotations, \footnotemark[3]\footnotetext[3]{Similar to KITTI, VoD plans to provide an evaluation server for testing.} we instead also used the validation data set for evaluation. Due to this procedure, our results differ from those in the original VoD paper~\cite{ViewOfDelftDataset_22} since they evaluate their models on their test set. VoD is highly focused on low-speed inner-city traffic scenes where most of the space is shared between all three types of traffic participants: pedestrians, cyclists, and vehicles. We used a data configuration where the point clouds of five consecutive frames have been aggregated (while maintaining temporal information) to one sample to increase the point cloud density. A detailed comparison between the VoD data set, released recently and specifically intended to stimulate research on 3D high-resolution radar perception, and other commonly established radar data sets (mostly not particularly meant for 3D object detection) is provided in the supplementary material. This overview is slightly more specific than the one in~\cite{RadarSurvey_22} in some practical details. \textbf{Astyx} is used to evaluate the detectors on another data set in a different environment. It mainly contains out-of-town industrial area environments where vehicles are by far the most frequent traffic participants. Only the radar point cloud is considered (lidar is rather sparse due to its 16 layers). A sample also only contains radar measurements from a one frame since the frames are not consecutive and hence cannot be aggregated, resulting in a sparser point cloud compared to VoD (Astyx contains roughly as many points per frame as VoD in its unaggregated version). The Astyx data set is very small, containing only 546 samples. Therefore, we used VoD pre-trained models and fine-tuned them for up to 50 epochs on 200 samples. The remaining 346 samples were used for evaluation.

\textbf{Evaluation Metrics} \hspace{0.25cm} As introduced by the KITTI data set~\cite{KITTIDataset_13} and used by many other researchers, the class-wise Average Precision (AP) and the mean AP (mAP) for a certain Intersection over Union (IoU) are used as the primary evaluation criteria. The evaluation distinguishes between 3D and 2D BEV. The precision-recall curve is evaluated at 40 sampling points, as for KITTI.

For all experiments, we conduct three runs with different and reproducible initializations of the models. Hence, we specify the mean and the standard deviation, which can indicate the robustness of the detectors to their initialization.

\begin{table*}
	\centering
	\caption{Evaluation of the reproducibility of the results from~\cite{ViewOfDelftDataset_22} and the influence of different IoU values for the pedestrian and cyclist class. As the text mentions, models marked by the prefix * are evaluated with stricter IoUs, while the other models use the IoU values from~\cite{ViewOfDelftDataset_22}.}
	\vspace{-0.125cm}
	\begin{tabular}{@{\extracolsep{1pt}}p{3.25cm}p{1.25cm}p{1.25cm} | p{1.25cm}p{1.25cm}p{1.25cm}p{1.25cm}p{1.25cm}p{1.25cm}}
		\toprule   
		{} & \multicolumn{2}{c|}{mAP} & \multicolumn{2}{c}{Car} & \multicolumn{2}{c}{Pedestrian} & \multicolumn{2}{c}{Cyclist} \\
		\cmidrule{2-3}
		\cmidrule{4-5}
		\cmidrule{6-7}
		\cmidrule{8-9}
		Model & 3D & BEV & 3D & BEV & 3D & BEV & 3D & BEV \\ 
		\midrule
		PointPillars-L (VoD) & 62.1 & - & 76.5 & - & 55.1 & - & 55.4 & - \\
		PointPillars-L (ours) & 65.9$\pm$0.6 & 67.7$\pm$1.3 & 66.6$\pm$0.4 & 71.7$\pm$2.4 & 56.1$\pm$0.5 & 56.3$\pm$0.5 & 75.1$\pm$1.0 & 75.1$\pm$1.0 \\
		\midrule
		PointPillars-R (VoD) & 47.0 & - & 44.8 & - & 42.1 & - & 54.0 & - \\
		PointPillars-R (ours) & 45.5$\pm$1.9 & 52.7$\pm$1.6 & 39.4$\pm$0.6 & 48.4$\pm$3.8 & 32.7$\pm$2.6 & 40.5$\pm$3.7 & 65.6$\pm$1.4 & 67.4$\pm$0.3  \\
		\midrule
		\midrule
		*PointPillars-L (ours) & 60.3$\pm$0.9 & 64.8$\pm$1.6 & 66.6$\pm$0.4 & 71.7$\pm$2.4 & 41.9$\pm$0.9 & 49.9$\pm$0.9 & 72.3$\pm$1.4 & 73.8$\pm$1.6  \\
		\midrule
		*PointPillars-R (ours) & 29.6$\pm$0.6 & 42.0$\pm$1.8 & 39.4$\pm$0.6 & 48.4$\pm$3.8 & 13.7$\pm$0.1 & 22.2$\pm$1.4 & 35.7$\pm$1.0 & 55.6$\pm$0.2  \\
		\bottomrule
	\end{tabular}
	\label{tab:comparison_VoD_paper}
\end{table*}

\subsection{Results and Discussion} \label{subsec:results_and_discussion}
\textbf{Comparison to VoD~\cite{ViewOfDelftDataset_22}} \hspace{0.25cm} First, we repeat the experiments and evaluation of the original PointPillars model~\cite{Pointpillar} and the adapted PointPillars-R model~\cite{ViewOfDelftDataset_22} on the VoD data set. Despite trying out further modifications to the provided code, we could not reproduce the results reported in~\cite{ViewOfDelftDataset_22}. From \cref{tab:comparison_VoD_paper} one can see that significant differences in the results remain for lidar and radar data. Our results are consistently better for the cyclist class but do not allow a clear ranking for the car and pedestrian classes\footnotemark[4]\footnotetext[4]{Such observations have been made by other researchers, too, according to the issues of the VoD Git repository \url{https://github.com/tudelft-iv/view-of-delft-dataset/issues}.}. Overall, our results on lidar are a bit better than those reported in~\cite{ViewOfDelftDataset_22}, while the results on radar are marginally worse than those mentioned in the paper. One reason could be that our test set was chosen to be different from the VoD one, as explained before. Hence, the composition of the evaluation data could be different concerning the distribution of object classes and the difficulty of the respective objects.

\begin{table*}
	\centering
	\caption{3D object detection results on the VoD lidar data. The best results are marked in \textbf{bold} font. $^{\dagger}$ marks one-stage detectors. This highlighting is used for the following tables, too. For clarity reasons, we omit to specify the standard deviation in the paper itself from now on. Instead, we duplicate the result tables in the supplementary material and also state the standard deviations there.}
	\vspace{-0.125cm}
	\begin{tabular}{@{\extracolsep{1pt}}p{2.25cm}p{0.475cm}p{0.475cm}p{0.475cm}p{0.475cm} | p{0.475cm}p{0.475cm}p{0.475cm}p{0.475cm}p{0.475cm}p{0.475cm}p{0.475cm}p{0.475cm}p{0.475cm}p{0.475cm}p{0.475cm}p{0.475cm}}
		\toprule   
		{} & \multicolumn{4}{c|}{mAP} & \multicolumn{4}{c}{Car} & \multicolumn{4}{c}{Pedestrian} & \multicolumn{4}{c}{Cyclist} \\
		\cmidrule{2-5}
		\cmidrule{6-9}
		\cmidrule{10-13}
		\cmidrule{14-17}
		{} & \multicolumn{2}{c}{3D} & \multicolumn{2}{c|}{BEV} & \multicolumn{2}{c}{3D} & \multicolumn{2}{c}{BEV} & \multicolumn{2}{c}{3D} & \multicolumn{2}{c}{BEV} & \multicolumn{2}{c}{3D} & \multicolumn{2}{c}{BEV} \\
		\cmidrule{2-3}
		\cmidrule{4-5}
		\cmidrule{6-7}
		\cmidrule{8-9}
		\cmidrule{10-11}
		\cmidrule{12-13}
		\cmidrule{14-15}
		\cmidrule{16-17}
		Model & SR & MR & SR & MR & SR & MR & SR & MR & SR & MR & SR & MR & SR & MR & SR & MR \\ 
		\midrule
		3DSSD$^{\dagger}$ & 72.5 & 53.5 & 73.1 & 57.1 & \textbf{81.0} & \textbf{69.1} & \textbf{81.3} & \textbf{76.1} & 57.5 & 38.8 & 59.1 & 42.5 & 79.0 & 52.6 & 79.0 & 52.8 \\
		Point-RCNN & \textbf{77.7} & 45.6 & \textbf{78.9} & 47.7 & \textbf{81.0} & 43.0 & 81.1 & 45.5 & \textbf{69.8} & 35.8 & \textbf{70.5} & 39.4 & 82.5 & 58.1 & 85.2 & 58.2 \\
		\midrule
		SECOND$^{\dagger}$ & 72.3 & 55.2 & 75.8 & 59.8 & 72.5 & 65.0 & 74.6 & 69.6 & 61.9 & 42.2 & 68.3 & 50.2 & 82.4 & \textbf{58.4} & 84.3 & 59.6 \\
		SECOND-MH$^{\dagger}$ & 73.8 & 55.6 & 74.7 & \textbf{60.8} & 72.5 & 65.5 & 74.5 & 69.7 & 65.5 & \textbf{43.9} & 65.7 & \textbf{52.7} & 83.3 & 57.3 & 84.0 & \textbf{60.0} \\
		SECOND-IoU$^{\dagger}$ & 71.7 & 51.8 & 75.0 & 57.7 & 72.4 & 62.9 & 74.4 & 69.8 & 61.5 & 39.3 & 67.2 & 46.7 & 81.2 & 53.3 & 83.4 & 56.6 \\
		Part A\textsuperscript{2} & 73.7 & 53.1 & 76.2 & 57.9 & 72.5 & 61.3 & 74.4 & 66.9 & 67.0 & 41.7 & 68.9 & 48.2 & 81.4 & 56.4 & 85.4 & 58.5 \\
		Voxel R-CNN & 74.7 & \textbf{55.7} & 75.3 & 58.6 & 72.0 & 66.5 & 72.2 & 68.8 & 66.7 & 42.4 & 68.2 & 47.8 & \textbf{85.5} & 58.2 & \textbf{85.6} & 59.2 \\
		\midrule
		PointPillars-L$^{\dagger}$ & 65.4 & 48.3 & 68.5 & 55.5 & 71.5 & 60.2 & 75.6 & 68.5 & 46.2 & 33.3 & 51.4 & 43.5 & 78.5 & 51.5 & 78.6 & 54.5 \\
		CenterPoint-L & 66.9 & 49.5 & 69.5 & 55.8 & 71.2 & 58.9 & 72.0 & 67.8 & 50.4 & 38.4 & 55.3 & 46.3 & 79.3 & 51.1 & 81.2 & 53.4 \\
		\midrule
		PV-RCNN & 71.9 & 53.3 & 75.3 & 58.6 & 76.1 & 65.6 & 76.1 & 65.6 & 63.3 & 41.2 & 65.3 & 47.1 & 80.8 & 57.5 & 80.9 & 58.4 \\
		\bottomrule
	\end{tabular}
	\label{tab:comparison_lidar}
	\vspace{-0.25cm}
\end{table*}

For the lidar results in \cref{tab:comparison_VoD_paper}, it can be observed that the differences between the 3D and BEV results are small or even zero for pedestrians and cyclists. Therefore, we also evaluated our trained models with a larger IoU than the ones used for both classes by~\cite{ViewOfDelftDataset_22}. In the lower part of the table, both trained models (lidar and radar) are evaluated with IoUs of $(0.5,\,0.5,\,0.5)$ for pedestrians and cyclists. That leads to an increased difference between the 3D and BEV results.

\textbf{Cross-Model Evaluation on Lidar Data} \hspace{0.25cm} Next, we do a cross-model validation on the VoD lidar data. This benchmark is novel since models other than PointPillars have not yet been evaluated on the VoD data set. The results in \cref{tab:comparison_lidar} are presented to enable the estimation of the influence of the data set in contrast to the effect of the respective detector models. Specifically, due to the different composition (ratio of object classes and different average velocities) of the data set, observations and properties for models listed in \cref{sec:state-of-the-art} for other data sets may not hold for VoD. Analogous to the evaluation of the Waymo data set~\cite{WaymoDataset_20} in~\cite{LidarSurvey_21}, we specify the results for three distances too: short-range (SR): \SI{0}{}-\SI{30}{\meter}, mid-range (MR): \SI{30}{}-\SI{50}{\meter}, and long-range (LR): $>$\SI{50}{\meter}. VoD does not contain annotations beyond \SI{51.2}{\meter}, preventing the evaluation at LR.

\begin{figure}
	\centering
	\includegraphics[width=0.5\textwidth,trim={0.5cm 0.5cm 0 0},clip]{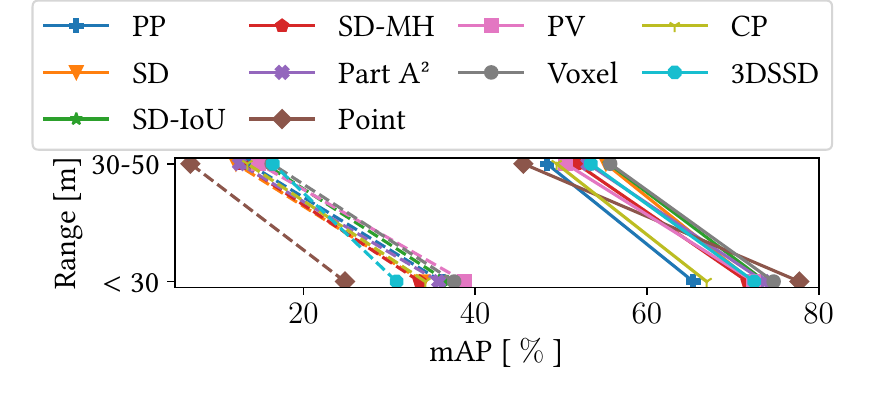}
	\caption{General trends for the detection performances of all the investigated models for both ranges on lidar (solid lines) and radar (dashed lines) data. PP: PointPillars, SD: SECOND, Point: Point-RCNN, PV: PV-RCNN, Voxel: Voxel R-CNN, CP: CenterPoint}
	\label{fig:Performance_VR}
	\vspace{-0.375cm}
\end{figure}

General trends in the lidar (and radar) detection results, represented by the mean average precision with respect to range, are summarized in \cref{fig:Performance_VR}. As can be seen in \cref{tab:comparison_lidar}, Point-RCNN yields good results in short-range but degrades in mid-range. The pillar-based models are inferior to PV-RCNN and the voxel-based models, which show robust and good detection results. 3DSSD excels in the car class and reaches similar results as the voxel-based models in general.

When comparing our results against the numbers in~\cite{LidarSurvey_21} (p.~24) for the Waymo data set, it has to be considered that this data set~\cite{WaymoDataset_20} contains about 140 times more cars, 60 times more pedestrians, but only twice as many cyclists as VoD. This explains why our detection results on VoD in \cref{tab:comparison_lidar} are inferior to those on the Waymo data set in~\cite{LidarSurvey_21}.

\begin{table*}
	\centering
	\caption{3D object detection results on the VoD radar data.}
	\vspace{-0.125cm}
	\begin{tabular}{@{\extracolsep{1pt}}p{2.25cm}p{0.475cm}p{0.475cm}p{0.475cm}p{0.475cm} | p{0.475cm}p{0.475cm}p{0.475cm}p{0.475cm}p{0.475cm}p{0.475cm}p{0.475cm}p{0.475cm}p{0.475cm}p{0.475cm}p{0.475cm}p{0.475cm}}
		\toprule   
		{} & \multicolumn{4}{c|}{mAP} & \multicolumn{4}{c}{Car} & \multicolumn{4}{c}{Pedestrian} & \multicolumn{4}{c}{Cyclist} \\
		\cmidrule{2-5}
		\cmidrule{6-9}
		\cmidrule{10-13}
		\cmidrule{14-17}
		{} & \multicolumn{2}{c}{3D} & \multicolumn{2}{c|}{BEV} & \multicolumn{2}{c}{3D} & \multicolumn{2}{c}{BEV} & \multicolumn{2}{c}{3D} & \multicolumn{2}{c}{BEV} & \multicolumn{2}{c}{3D} & \multicolumn{2}{c}{BEV} \\
		\cmidrule{2-3}
		\cmidrule{4-5}
		\cmidrule{6-7}
		\cmidrule{8-9}
		\cmidrule{10-11}
		\cmidrule{12-13}
		\cmidrule{14-15}
		\cmidrule{16-17}
		Model & SR & MR & SR & MR & SR & MR & SR & MR & SR & MR & SR & MR & SR & MR & SR & MR \\ 
		\midrule
		3DSSD$^{\dagger}$ & 30.8 & \textbf{16.4} & 38.7 & 24.4 & 46.5 & \textbf{27.2} & 49.9 & 39.2 & 11.9 & 7.6 & 19.1 & 10.9 & 34.1 & 14.5 & 47.2 & 23.1 \\
		Point-RCNN & 26.5 & 7.1 & 36.9 & 10.7 & 28.8 & 9.1 & 32.5 & 11.2 & 20.3 & 6.6 & 29.6 & 8.2 & 30.5 & 6.1 & 48.5 & 12.7 \\
		\midrule
		SECOND$^{\dagger}$ & 33.9 & 12.1 & 45.5 & 20.7 & 45.5 & 20.8 & 51.4 & 30.1 & 18.0 & 5.8 & 27.6 & 11.6 & 38.3 & 9.8 & 57.5 & 20.3 \\
		SECOND-MH$^{\dagger}$ & 36.8 & 15.5 & 42.8 & 24.0 & \textbf{47.9} & 22.6 & \textbf{52.1} & 32.7 & 19.9 & 6.0 & 25.3 & 12.5 & 42.5 & 17.8 & 51.0 & 26.9 \\
		SECOND-IoU$^{\dagger}$ & 33.5 & 12.7 & 42.6 & 21.6 & 47.6 & 21.8 & 51.4 & 32.5 & 17.0 & 2.0 & 25.3 & 5.2 & 36.0 & 14.4 & 51.0 & 27.3 \\
		Part A\textsuperscript{2} & 35.7 & 12.5 & 43.7 & 18.8 & 43.2 & 17.9 & 44.3 & 22.9 & 21.5 & 5.5 & \textbf{29.9} & 8.8 & 42.5 & 14.2 & 56.8  & 24.6 \\
		Voxel R-CNN & 37.5 & 16.3 & 43.0 & 26.8 & 44.7 & 22.8 & 49.7 & 30.3 & \textbf{24.2} & \textbf{7.7} & 25.9 & \textbf{19.4} & 43.7 & 18.4 & 53.5 & 30.7 \\
		\midrule
		PointPillars-R$^{\dagger}$ & 36.1 & 13.6 & \textbf{48.1} & 28.4 & 46.1 & 27.1 & 51.7 & \textbf{45.5} & 16.5 & 1.7 & 26.9 & 5.8 & 45.7 & 11.9 & \textbf{65.7} & 34.0 \\
		CenterPoint-R & 34.2 & 13.5 & 46.2 & 24.5 & 43.6 & 21.5 & 47.1 & 35.0 & 19.1 & 2.2 & 29.5 & 8.6 & 39.8 & 16.6 & 62.1 & 30.0 \\
		\midrule
		PV-RCNN & \textbf{38.8} & 14.8 & 44.6 & \textbf{32.4} & 45.2 & 22.9 & 46.7 & 36.6 & 21.8 & 2.3 & 27.7  & 16.2 & \textbf{49.3} & \textbf{19.1} & 59.5 & \textbf{44.4} \\
		\bottomrule
	\end{tabular}
	\label{tab:comparison_radar}
	\vspace{-0.25cm}
\end{table*}

\textbf{Cross-Model Evaluation on Radar Data} \hspace{0.25cm} \cref{tab:comparison_radar} shows the results of the numerical study for the VoD radar data. While Point-RCNN performs significantly worse than all the other models, 3DSSD's performance clearly declines less. There is only a slight gap between SECOND, SECOND-IoU, CenterPoint, and the remaining models. The adapted PointPillars-R model~\cite{ViewOfDelftDataset_22} is among the best detectors. Overall, there is no clear best-performing model class.

\begin{table}
	\centering
	\caption{Object detection results on the Astyx data for the car class only. Since other object classes are rare, only this class is evaluated.}
	\vspace{-0.125cm}
	\begin{tabular}{@{\extracolsep{1pt}}p{2.25cm}p{0.475cm}p{0.475cm}p{0.475cm}p{0.475cm}p{0.475cm}p{0.475cm}}
		\toprule   
		{} & \multicolumn{3}{c}{3D} & \multicolumn{3}{c}{BEV} \\
		\cmidrule{2-4}
		\cmidrule{5-7}
		Model & SR & MR & LR & SR & MR & LR \\ 
		\midrule
		3DSSD$^{\dagger}$ & 17.5 & 4.6 & 3.6 & 34.1 & 14.7 & 7.1 \\
		Point-RCNN & 2.5 & 0.4 & 0.2 & 8.7 & 3.0 & 0.3 \\
		\midrule
		SECOND$^{\dagger}$ & 13.0 & 6.1 & 1.1 & 25.0 & 19.4 & 12.3 \\
		SECOND-MH$^{\dagger}$ & 19.7 & 9.6 & 2.6 & 40.2 & 24.6 & 15.2 \\
		SECOND-IoU$^{\dagger}$ & 20.1 & 8.3 & 4.2 & 35.1 & 25.4  & \textbf{16.4} \\
		Part A\textsuperscript{2} & 9.9  & 2.1 & 1.0 & 19.9 & 7.5 & 6.0 \\
		Voxel R-CNN & 20.9 & 6.4 &  1.4 &  38.7 &  21.0 &  11.4 \\
		\midrule
		PointPillars-R$^{\dagger}$ & 14.1 & 2.2 & 0.2 & \textbf{40.8} & 22.2 & 13.9 \\
		CenterPoint-R & 22.6 & \textbf{10.4} & \textbf{6.1} & 37.6 & 21.9 & 9.2 \\
		\midrule
		PV-RCNN & \textbf{24.4} & 9.0 & 3.0 & 39.8 & \textbf{26.7} & 15.4 \\
		\bottomrule
	\end{tabular}
	\label{tab:astyx_evaluation}
	\vspace{-0.5cm}
\end{table}

\textbf{Cross-Model Evaluation on Astyx Data} \hspace{0.25cm} As a next step, the radar detection models previously trained on VoD are fine-tuned on Astyx to account for different sensor characteristics and the shifted data distribution. The results on this second data set are reported in \cref{tab:astyx_evaluation}. Note, different factors might cause the generally worse accuracy. First, as mentioned in \cref{subsec:experimental_setup}, Astyx point clouds are, on average one-fifth sparser than VoD point clouds. The second influence limiting the detection performance is the data set size. The 200 samples used for fine-tuning capture only a limited diversity. Training on such a small data set prevents the model from reaching good generalization performance. However, trends observed on the VoD radar data set before could be confirmed on Astyx. Voxel R-CNN, CenterPoint, and PV-RCNN are among the best-performing models. 

\begin{table*}
	\centering
	\caption{Investigation of small voxel and pillar sizes (svs, sps) vs. large voxel and pillar sizes (lvs, lps). For the adapted SECOND model with lvs we additionally experimented with an adapted learning rate scheduler due to the insides from the training of the initial model.}
	\vspace{-0.125cm}
	\begin{tabular}{@{\extracolsep{1pt}}p{2.125cm}p{0.475cm}p{0.475cm}p{0.475cm}p{0.475cm} | p{0.475cm}p{0.475cm}p{0.475cm}p{0.475cm}p{0.475cm}p{0.475cm}p{0.475cm}p{0.475cm}p{0.475cm}p{0.475cm}p{0.475cm}p{0.475cm}}
		\toprule   
		{} & \multicolumn{4}{c|}{mAP} & \multicolumn{4}{c}{Car} & \multicolumn{4}{c}{Pedestrian} & \multicolumn{4}{c}{Cyclist} \\
		\cmidrule{2-5}
		\cmidrule{6-9}
		\cmidrule{10-13}
		\cmidrule{14-17}
		{} & \multicolumn{2}{c}{3D} & \multicolumn{2}{c|}{BEV} & \multicolumn{2}{c}{3D} & \multicolumn{2}{c}{BEV} & \multicolumn{2}{c}{3D} & \multicolumn{2}{c}{BEV} & \multicolumn{2}{c}{3D} & \multicolumn{2}{c}{BEV} \\
		\cmidrule{2-3}
		\cmidrule{4-5}
		\cmidrule{6-7}
		\cmidrule{8-9}
		\cmidrule{10-11}
		\cmidrule{12-13}
		\cmidrule{14-15}
		\cmidrule{16-17}
		Model & SR & MR & SR & MR & SR & MR & SR & MR & SR & MR & SR & MR & SR & MR & SR & MR \\ 
		\midrule
		PointPillars-R & \textbf{36.1} & \textbf{13.6} & 48.1 & \textbf{28.4} & 46.1 & \textbf{27.1} & \textbf{51.7} & \textbf{45.5} & 16.5 & 1.7 & 26.9 & 5.8 & \textbf{45.7} & \textbf{11.9} & 65.7 & \textbf{34.0} \\
		\midrule
		PointPillars-R (sps) & 35.5 & 11.4 & \textbf{50.1} & 27.2 & \textbf{46.5} & 21.2 & 51.0 & 39.1 & \textbf{18.0} & \textbf{4.5} & \textbf{30.2} & \textbf{10.3} & 42.0 & 8.5 & \textbf{69.1} & 32.1 \\
		\midrule
		\midrule
		SECOND & \textbf{33.9} & \textbf{12.1} & \textbf{45.5} & \textbf{20.7} & \textbf{45.5} & \textbf{20.8} & \textbf{51.4} & \textbf{30.1} & \textbf{18.0} & 5.8 & \textbf{27.6} & \textbf{11.6} & 38.3 & \textbf{9.8} & \textbf{57.5} & \textbf{20.3} \\
		\midrule
		SECOND (lvs) & 29.8 & 10.4 & 35.8 & 17.5 & 43.6 & 18.0 & 46.0 & 27.3 & 11.9 & \textbf{6.7} & 15.7 & 8.1 & 33.8 & 6.5 & 45.7 & 17.0 \\
		SECOND (lvs, lr scheduler) & 32.4 & 11.2 & 38.9 & 17.6 & 44.3 & 17.3 & 45.0 & 26.5 & 13.1 & 7.1 & 17.9 & 9.1 & \textbf{39.8} & 9.3 & 53.9 & 17.1 \\
		\bottomrule
	\end{tabular}
	\label{tab:detailed_evaluation_second}
	\vspace{-0.25cm}
\end{table*}

\textbf{Influence of Pillar and Voxel Sizes} \hspace{0.25cm} The authors of~\cite{ViewOfDelftDataset_22} indicate a slight adaptation of the pillar height from \SI{4}{\meter} to \SI{5}{\meter} for their PointPillars-R model\footnotemark[2] with respect to the original OpenPCDet implementation. This motivates a more extensive investigation of the influence of smaller and larger pillar and voxel sizes due to the sparsity of the radar data. Larger volumetric units are supposed to capture coarser structures in the data supporting the detection of objects only represented by a few radar measurements. In particular, we investigate several adaptations of two models, PointPillars-R and SECOND. We chose PointPillars since it is used throughout the paper as a kind of reference, \eg, PointPillars-R~\cite{ViewOfDelftDataset_22}. SECOND has been selected since it is the weakest voxel-based detector on radar data, according to \cref{tab:comparison_radar}. Thus, adaptations might be most beneficial for this model. Since the pillars (in the $x-y$ plane) in PointPillars are already quite large (compared to the voxel size of voxel-based models), we scale them down to half of the initial value in x and y direction, resulting in a pillar dimension of $0.08 \times 0.08 \times 5.0$\SI{}{\meter}. Conversely, for SECOND we increased the default values voxel size from $0.036 \times 0.032 \times 0.125$\SI{}{\meter} used for lidar to $0.135 \times 0.120 \times 0.625$\SI{}{\meter}. The adapted voxel size was chosen to keep the modification to the original SECOND model as simple as possible. We explicitly only adapted the sparse 3D convolutional backbone to output a grid of the same size as the base model. While evaluating the adapted SECOND model, we adjusted the learning rate scheduler to extend the time for applying large lr.

\cref{tab:detailed_evaluation_second} shows the results for the original and the adapted models. In general, the original models perform better, except for PointPillars-R for the pedestrian class. Thus, the idea of aggregating more evidence for the noisy radar data by increasing the volume of the respective spatial unit (voxel and pillar) seems hardly verifiable by the numerical results. The numbers for the models are inconclusive and indicate that this design choice may be less important than assumed.

\textbf{Discussion} \hspace{0.25cm} There is a clear trend comparing the radar and lidar detection results. With increasing distance, the \textit{relative} gap between the detection accuracy widens. This behavior might seem counter-intuitive since the lidar point cloud density becomes sparser more quickly with increasing distance than the radar point cloud density, as seen in the supplement. Overall, the performance gap is significant for all object classes but not quantitatively equal. The detection performance suffers less for the cyclist class. Such behavior was described by~\cite{ViewOfDelftDataset_22} before and attributed to two reasons: first, the proportional number of moving objects is much higher for cyclists than for cars and pedestrians, and second, a moving bicycle has a high reflectivity due to its metal frame and highly reflective parts such as pedals. Another key difference between lidar and radar is the mounting positions of the sensors~\cite{ViewOfDelftDataset_22}. A lidar placed on the car's roof does not encounter as many occlusions as a radar mounted at the front bumper, which is the typical position of radar sensors. The vast majority of lidar points come from the ground. The lidar points reflected from non-ground objects are much denser than the accumulated radar point cloud only at close ranges, as seen in the supplementary material. However, the radar measurements are significantly noisier, resulting in targets outside of ground truth BEV rectangles as seen in the supplement. Radar measurements can also be significantly outside ground truth cuboids in vertical direction.
\newline $\Rightarrow$ Key finding 1: the performance gap cannot only be attributed to the radar's sparsity but also to its high noise level.

According to the results in \cref{tab:comparison_radar}, PV-RCNN and Voxel R-CNN perform slightly better than PointPillars-R concerning the mAP in 3D. Thus, additionally considering the detection results on the Astyx data from \cref{tab:astyx_evaluation}, PV-RCNN and Voxel R-CNN can be considered more robust over a wide range of different data configurations. However, there is no clear best-performing model. Only Point-RCNN can be considered unsuitable when applied to radar data. In general, different initializations significantly affect the performance (as can be seen in \cref{tab:comparison_VoD_paper}), additionally complicating the evaluation of the results (as mentioned before, the standard deviations for all other results are stated in the supplementary material). However, when additionally considering the qualitative results in the supplementary material, the error modes of the evaluated detectors on the radar data become obvious. First, several models suffer from many false positives (FPs). This is assumed to be caused by the sparsity of the radar point clouds. Many ground truth annotations only contain very few radar target points. Then, the detectors learn to generate detections even in sparse regions. Additionally, due to the high amount of noise in the data, the detected bounding boxes are often significantly off the ground truth annotations.
\newline $\Rightarrow$ Key finding 2: no clear best-performing model has been identified, but Point-RCNN is considered inferior.

While the sparsity cannot easily be resolved (without extending the aggregation horizon), the naive point cloud accumulation applied in VoD~\cite{ViewOfDelftDataset_22} could be improved. Radar sensors measure the relative radial velocity and aid the differentiation of static and dynamic objects~\cite{RadarSegmentation_18}. The estimated motion of dynamic objects can then be used to correct points before accumulation, which leads to a more consistent aggregated point cloud with fewer smearing artifacts of dynamic objects. This approach is supposed to improve object detection results. Alternatively, specific modules or strategies known to address the sparsity issue, such as self-attention~\cite{SADet3D} or additionally estimating the detection's uncertainty~\cite{3DUncertaintyEstimation_19}, could be applied to existing object detectors. Another approach is developing radar-specific architectures utilizing the measured relative radial velocity. Approaches like~\cite{DOPS_20} have been demonstrated to improve the detection performance in sparse regions by additionally completing object shapes.
\newline $\Rightarrow$ Possible solutions: correctly accumulating radar data improves detection, but radar-specific extensions are required to close the performance gap.

\textbf{Limitations} \hspace{0.25cm} The VoD data set prevents a general evaluation since it only contains inner-city traffic scenes, neglecting traffic situations involving higher velocities. The simple accumulation of dynamic objects is an additional limitation. The Astyx data set is very small and contains no sequential data. Also, we only adapted the first modules of the detectors to accept the additional radar inputs, \eg, relative radial velocity. Thus, we have not yet applied major architectural changes, particularly utilizing radar-specific inputs.

\section{Conclusion} \label{sec:conclusion}
We have investigated ten different 3D object detectors by evaluating their performance on radar data which were initially developed for lidar perception. However, as we have shown, the gap in the detection performance between radar and lidar perceptions remains significant even for the best-performing detectors. The numerical results also show that there is no clear best-performing model, but that results are mixed with respect to object class and object distance. More research along various avenues is required. First, a broader data base for 3+1D radar data explicitly intended for 3D object detection is required to increase the results' certainty. Secondly, since the results indicate that 3+1D radar-only object detection without further processing such as tracking, is insufficient for automated driving, previously described radar-specific modifications have to be applied.

%%%%%%%%% REFERENCES
{\small
\bibliographystyle{ieee_fullname}
\bibliography{cvpr_2023_references}
}

\clearpage

%%%%%%%%% APPENDIX
\appendix
\section{Appendix}
\subsection{Radar Data Set Comparison} \label{subsec:radar_dataset_comparison}
The plethora of currently available data sets with their specific characteristics containing radar measurements make it necessary to compare them in several dimensions. These dimensions should reflect different aspects crucial for the applicability of training a deep learning radar perception model that should support automated driving \cite{RadarNet_20, SeeingTroughFog_20, RadarTransformer_21}. We have focused on the following dimensions: the data set size, the radar sensor type, the area observed by the radar (field of view and view range), the data format, additional (reference) sensors, and available annotations. \cref{tab:datasets} gives an overview of the investigated data sets and their characteristics. Note that some data sets \cite{AstyxDataset_19, ColoradarDataset_19, MCityDataset_19, SCORPDataset_20, RaDICaLDataset_21} are not included in this comparison due to severe restrictions such as a too small radar view range or a too small size of the data set. The TJ4RadSet data set \cite{TJ4RadSet_22} is not available yet and hence not considered. On the other hand, K-Radar \cite{KRadarDataset_22} is already available but only as spectral data and not as point clouds, making it unsuitable for our purposes. While the Adverse Weather data set \cite{AdverseWeatherDataset_20} is rather large, it only contains spectral data and is thus not usable for us. Finally, MulRan \cite{MulRanDataset_20} is omitted since it is not intended for object detection and contains a spinning radar that does not measure radial velocity.

While \cite{RadarSurvey_22} also contains a comparison of different radar datasets, our focus is to be as specific as possible about technical details. For example, we not only categorize the used radar sensor in groups such as spinning, low-resolution, and high-resolution but specify exact resolution values for all relevant measurement dimensions, mention precise values for the view range and the field(s) of view (if available), and provide the sensor's brand and name. This lets practitioners decide if a specific data set is relevant for their task.

\subsection{Quantitative Results: Additional Standard Deviations} \label{subsec:standard_deviations}
The \cref{tab:comparison_lidar_appendix,tab:comparison_radar_appendix,tab:astyx_evaluation_appendix,tab:detailed_evaluation_second_appendix} extend \cref{tab:comparison_lidar,tab:comparison_radar,tab:astyx_evaluation,tab:detailed_evaluation_second} in the main paper by specifying the standard deviation for all APs and mAPs. One key observation from this additional information is the increased variation on the Astyx data set \cite{AstyxDataset_19} indicating a high sensitivity, which can partly be explained by the comparably low number of data points in the data set. Another significant characteristic is the increased standard deviation comparing mid-range and short-range results. However, such behavior is expected and reflects the rising difficulty of detecting objects in sparser point clouds.

\subsection{Qualitative Results} \label{subsec:qualitative_results}
\begin{figure*}[t!]
	\hfill
	\vspace{0.2cm}
	\begin{subfigure}{0.4\textwidth}
		\centering
		\includegraphics[width=\textwidth,trim={0 0 0 5.5cm},clip]{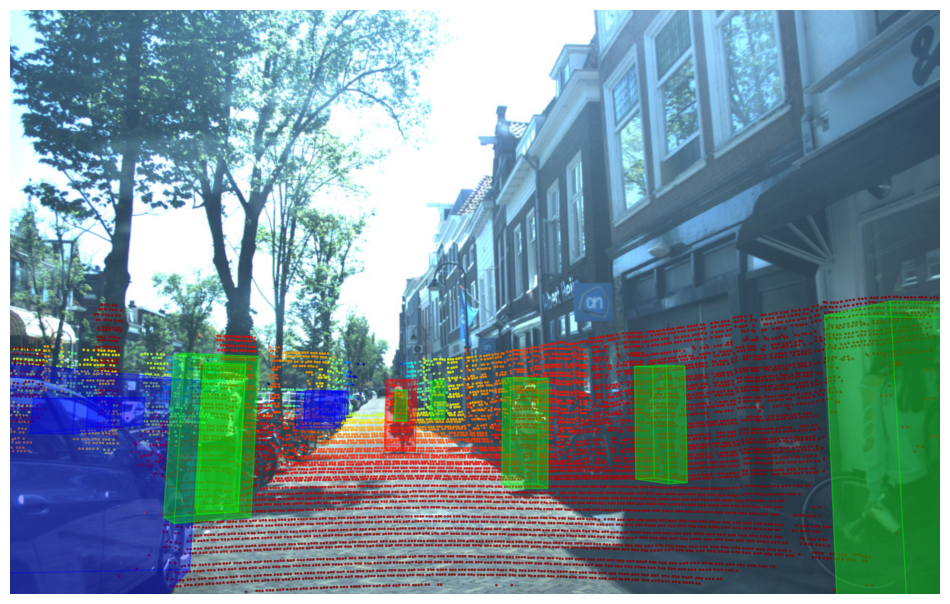}
		\caption{Ground truth bounding box annotations}
		\label{fig:gt_annotation_lidar}
	\end{subfigure}
	\hspace*{\fill}
	\\
	\hfill
	\vspace{0.2cm}
	\begin{subfigure}{0.4\textwidth}
		\centering
		\includegraphics[width=\textwidth,trim={0 0 0 5.5cm},clip]{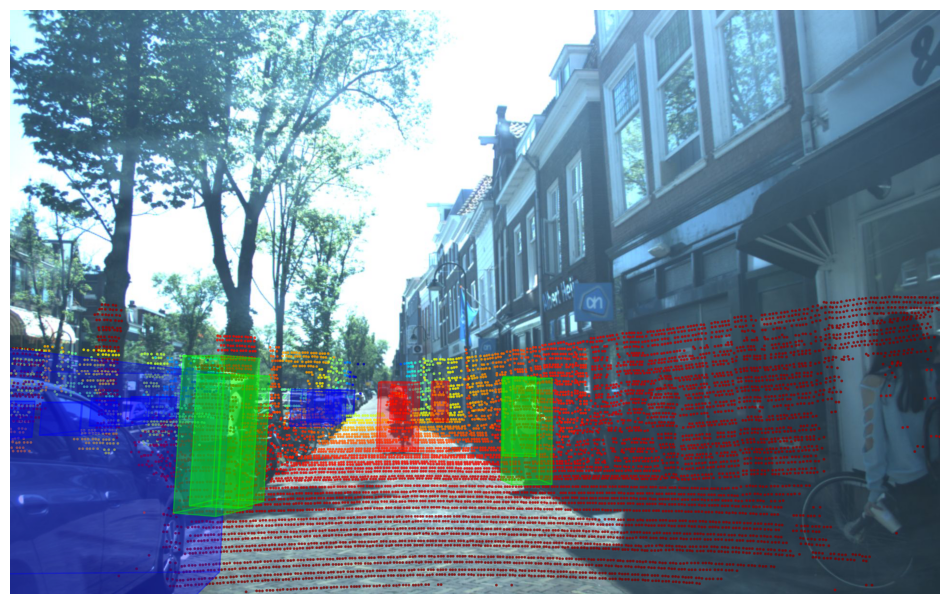}
		\caption{3DSSD}
		\label{fig:3dssd_lidar}
	\end{subfigure}
	\hfill
	\begin{subfigure}{0.4\textwidth}
		\centering
		\includegraphics[width=\textwidth,trim={0 0 0 5.5cm},clip]{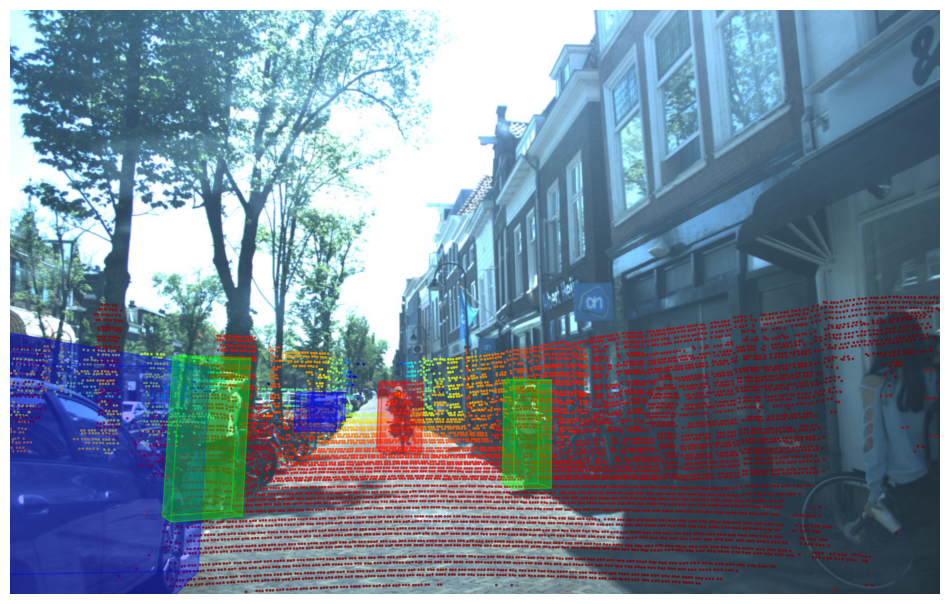}
		\caption{Point-RCNN}
		\label{fig:point-rcnn_lidar}
	\end{subfigure}
	\hfill
	\\
	\hfill
	\vspace{0.2cm}
	\begin{subfigure}{0.4\textwidth}
		\centering 
		\includegraphics[width=\textwidth,trim={0 0 0 5.5cm},clip]{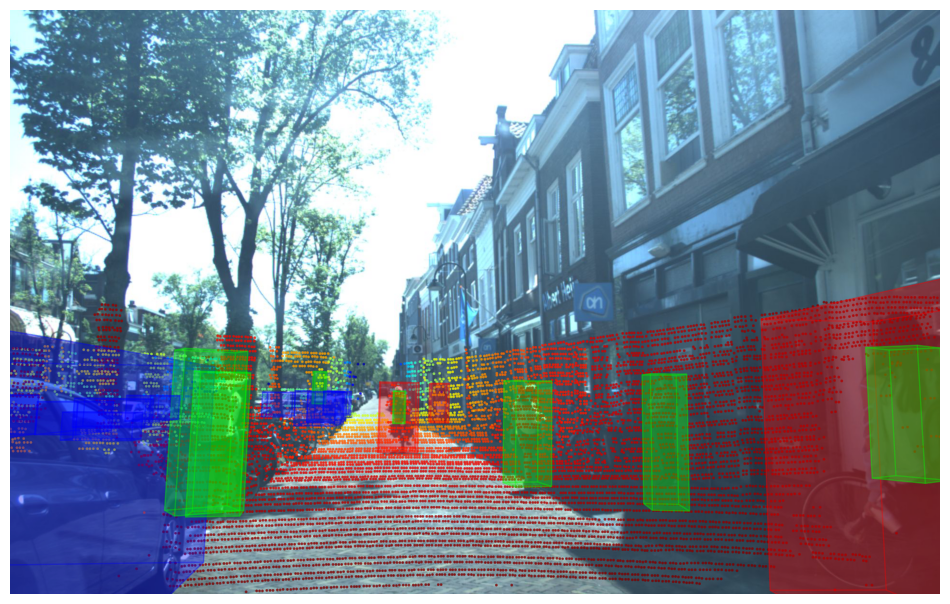}
		\caption{SECOND}
		\label{fig:second_lidar}
	\end{subfigure}
	\hfill
	\begin{subfigure}{0.4\textwidth}
		\centering 
		\includegraphics[width=\textwidth,trim={0 0 0 5.5cm},clip]{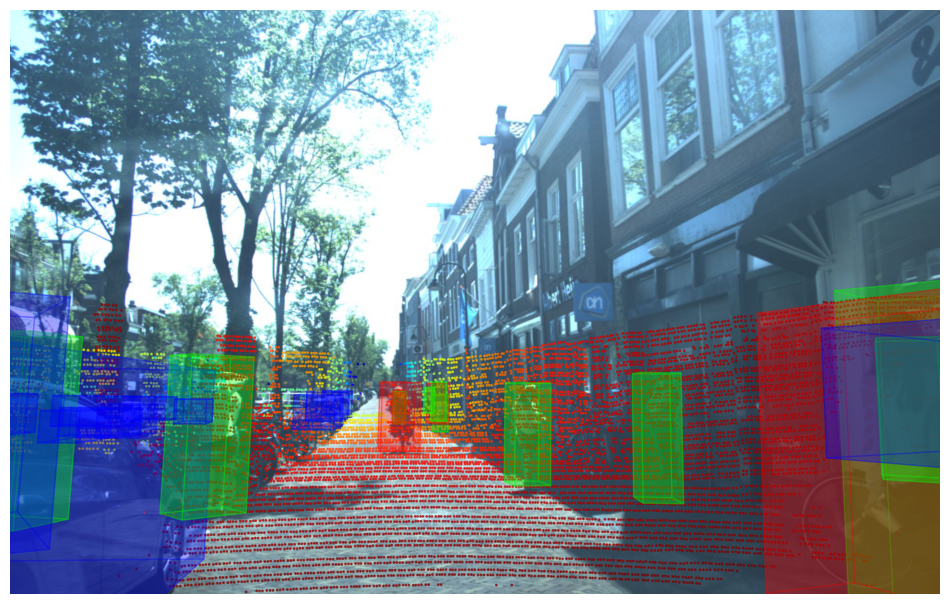}
		\caption{SECOND-MH}
		\label{fig:second-mh_lidar}
	\end{subfigure}
	\hfill
	\\
	\hfill
	\vspace{0.2cm}
	\begin{subfigure}{0.4\textwidth}
		\centering
		\includegraphics[width=\textwidth,trim={0 0 0 5.5cm},clip]{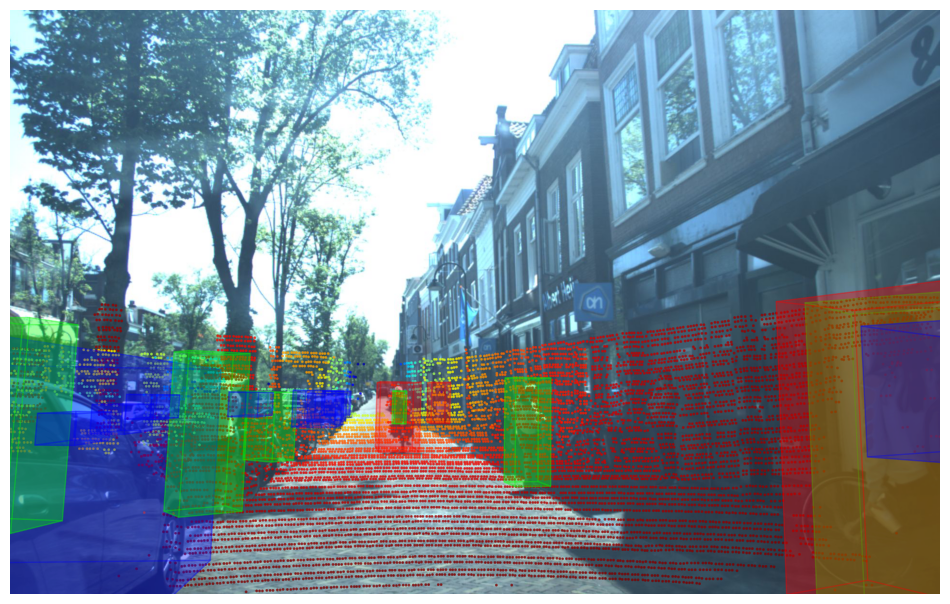}
		\caption{SECOND-IoU}
		\label{fig:second-iou_lidar}
	\end{subfigure}
	\hfill
	\begin{subfigure}{0.4\textwidth}
		\centering
		\includegraphics[width=\textwidth,trim={0 0 0 5.5cm},clip]{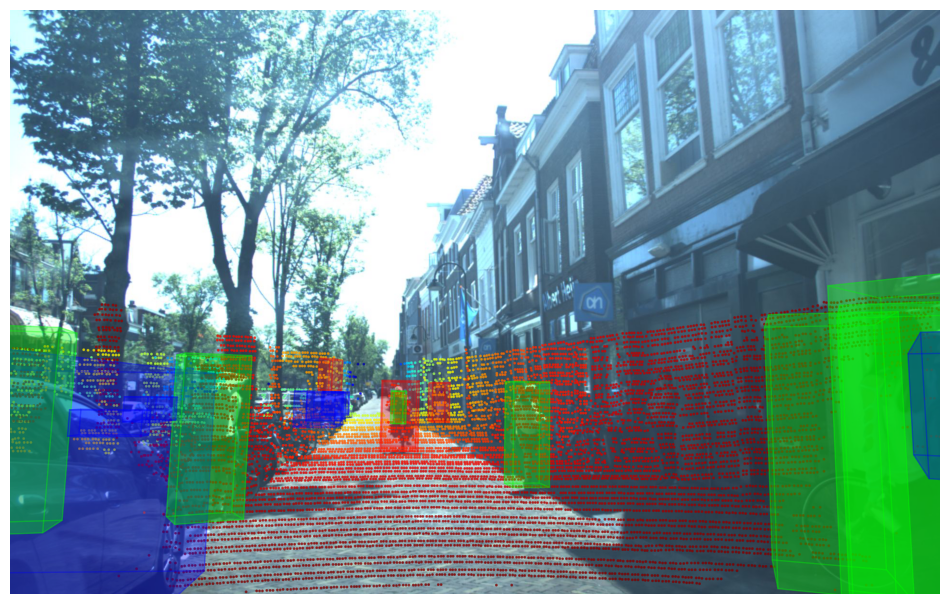}
		\caption{Part A\textsuperscript{2}}
		\label{fig:part_a2_lidar}
	\end{subfigure}
	\hfill
	\\
	\hfill
	\vspace{0.2cm}
	\begin{subfigure}{0.4\textwidth}
		\centering
		\includegraphics[width=\textwidth,trim={0 0 0 5.5cm},clip]{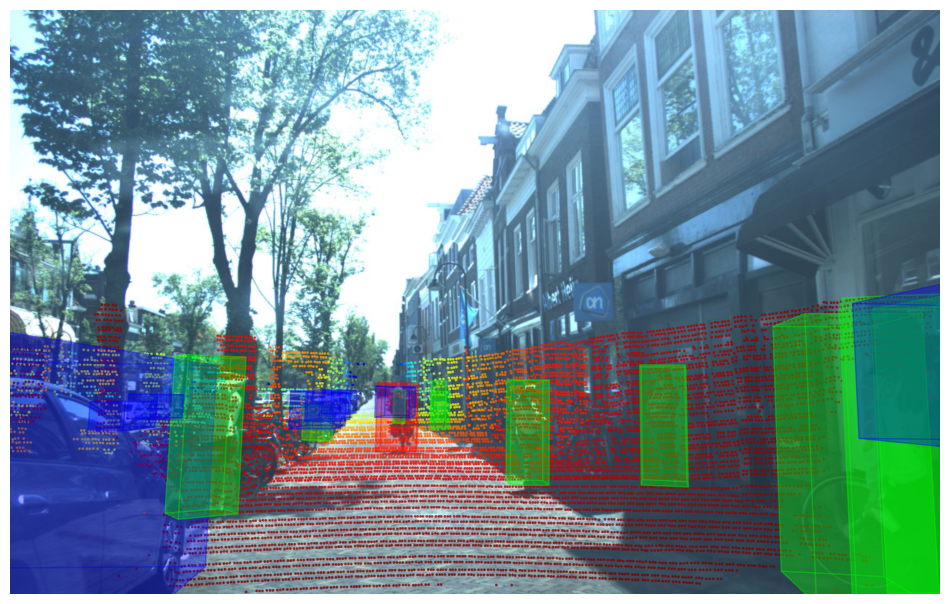}
		\caption{Voxel R-CNN}
		\label{fig:voxel-rcnn_lidar}
	\end{subfigure}	
	\hfill
	\begin{subfigure}{0.4\textwidth}
		\centering
		\includegraphics[width=\textwidth,trim={0 0 0 5.5cm},clip]{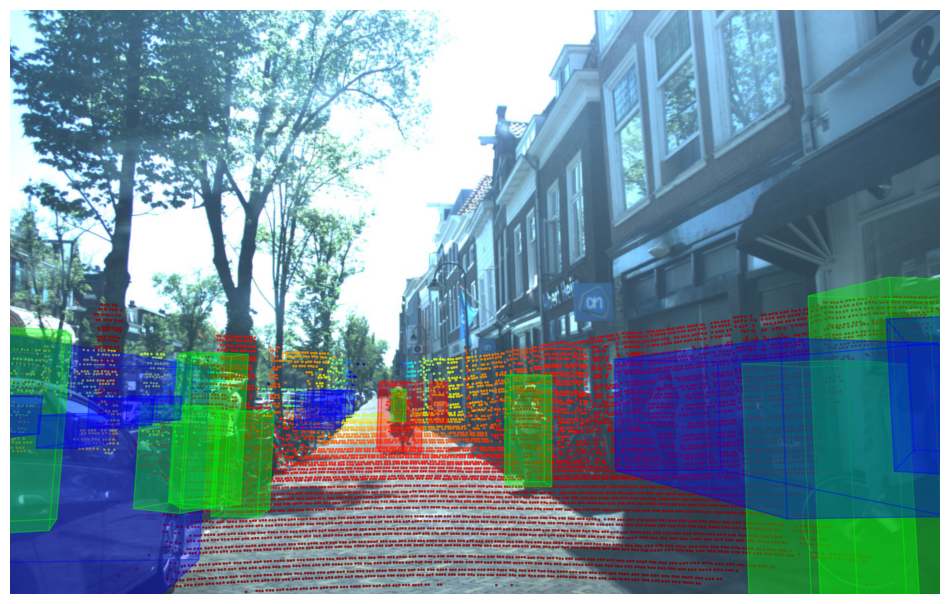}
		\caption{PointPillars-L}
		\label{fig:Point-RCNN}
	\end{subfigure}	
	\hfill
	\\
	\hfill
	\vspace{0.2cm}
	\begin{subfigure}{0.4\textwidth}
		\centering 
		\includegraphics[width=\textwidth,trim={0 0 0 5.5cm},clip]{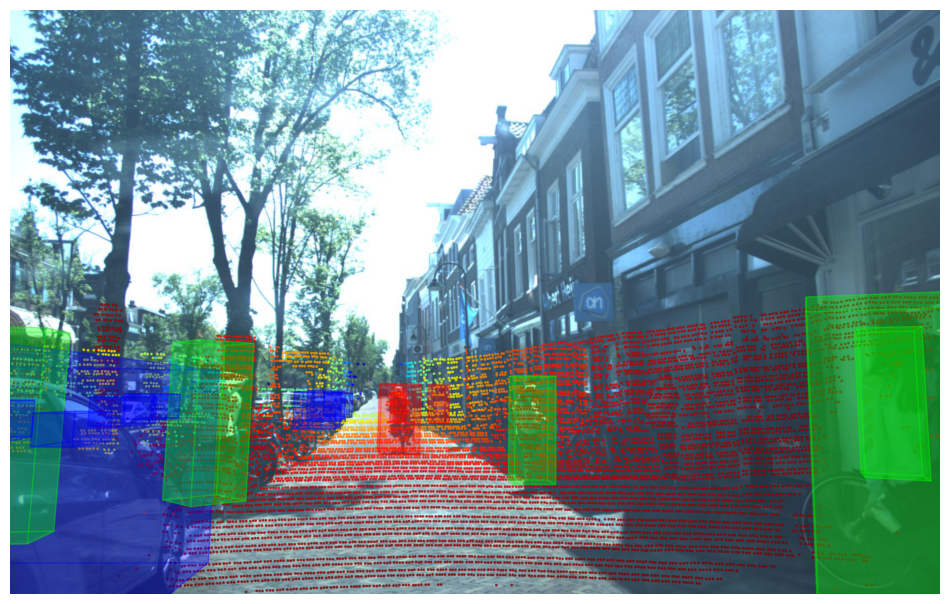}
		\caption{CenterPoint-L}
		\label{fig:Centerpoint-lidar}
	\end{subfigure}
	\hfill
	\begin{subfigure}{0.4\textwidth}
		\centering
		\includegraphics[width=\textwidth,trim={0 0 0 5.5cm},clip]{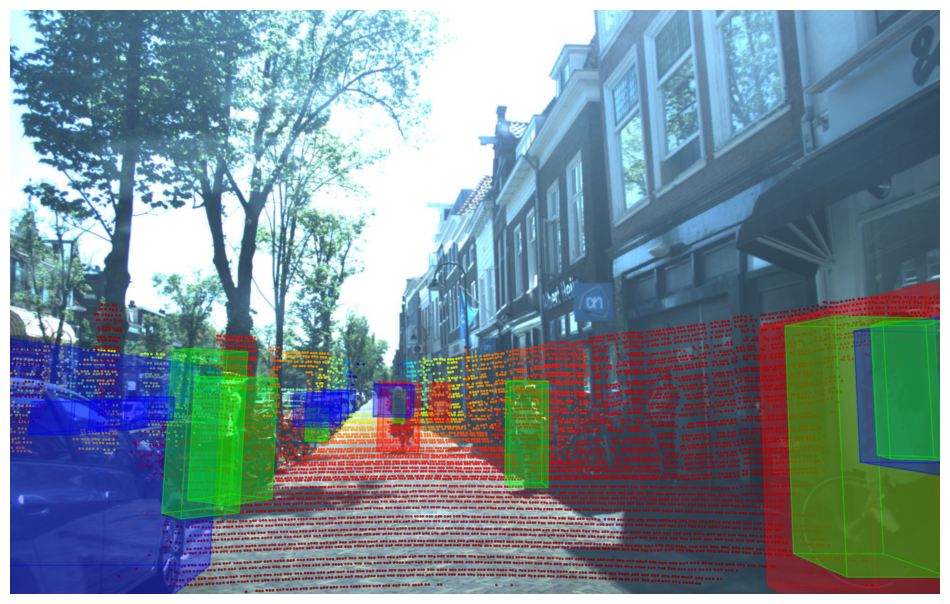}
		\caption{PV-RCNN}
		\label{fig:pv_rcnn}
	\end{subfigure}	
	\caption{Visualization of a scene from the VoD evaluation set. The camera image and the lidar point cloud are synchronized using the provided code. The colored boxes show the detections of the respective models and the ground truth annotations, respectively (blue: cars, red: cyclists, green: pedestrians). The camera image is cropped at the top to focus on the relevant image part containing objects.}
	\label{fig:lidar_detections}
\end{figure*}

\begin{figure*}
	\hfill
	\vspace{0.2cm}
	\begin{subfigure}{0.375\textwidth}
		\centering
		\includegraphics[width=\textwidth,trim={0 0 0 5.5cm},clip]{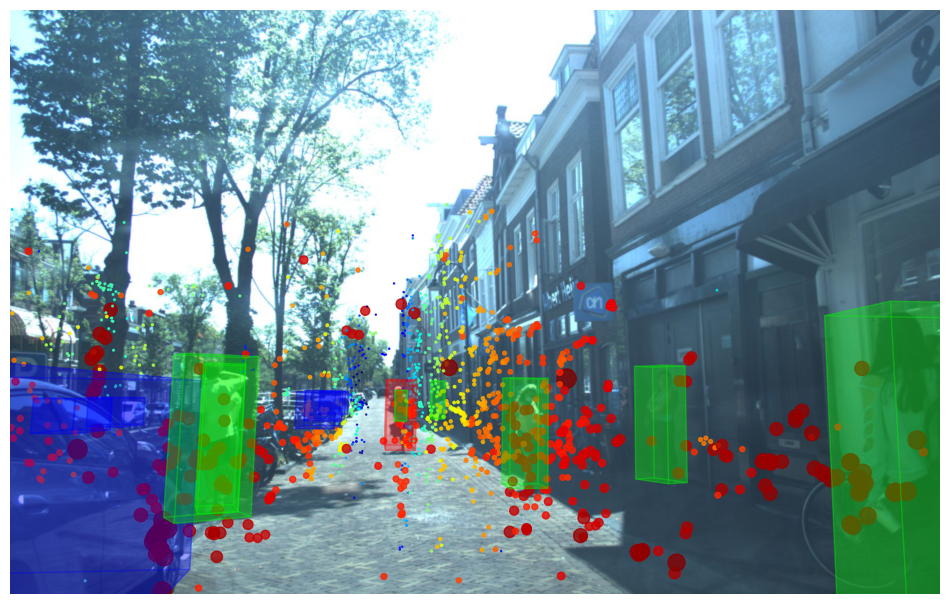}
		\caption{Ground truth bounding box annotations}
		\label{fig:gt_annotation}
	\end{subfigure}
	\hspace*{\fill}
	\\
	\hfill
	\vspace{0.2cm}
	\begin{subfigure}{0.375\textwidth}
		\centering
		\includegraphics[width=\textwidth,trim={0 0 0 5.5cm},clip]{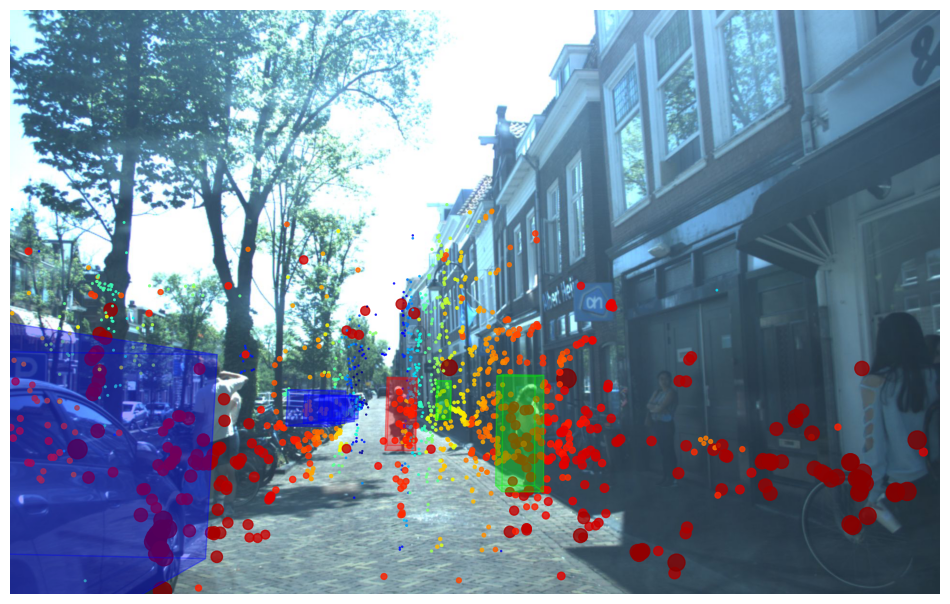}
		\caption{3DSSD}
		\label{fig:3dssd}
	\end{subfigure}
	\hfill
	\begin{subfigure}{0.375\textwidth}
		\centering
		\includegraphics[width=\textwidth,trim={0 0 0 5.5cm},clip]{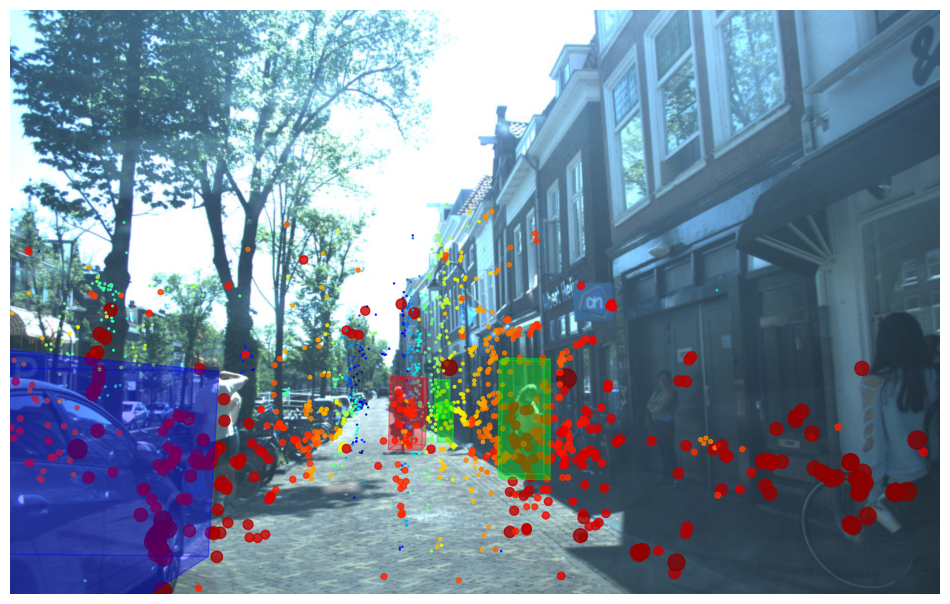}
		\caption{Point-RCNN}
		\label{fig:point-rcnn}
	\end{subfigure}
	\hfill
	\\
	\hfill
	\vspace{0.2cm}
	\begin{subfigure}{0.375\textwidth}
		\centering 
		\includegraphics[width=\textwidth,trim={0 0 0 5.5cm},clip]{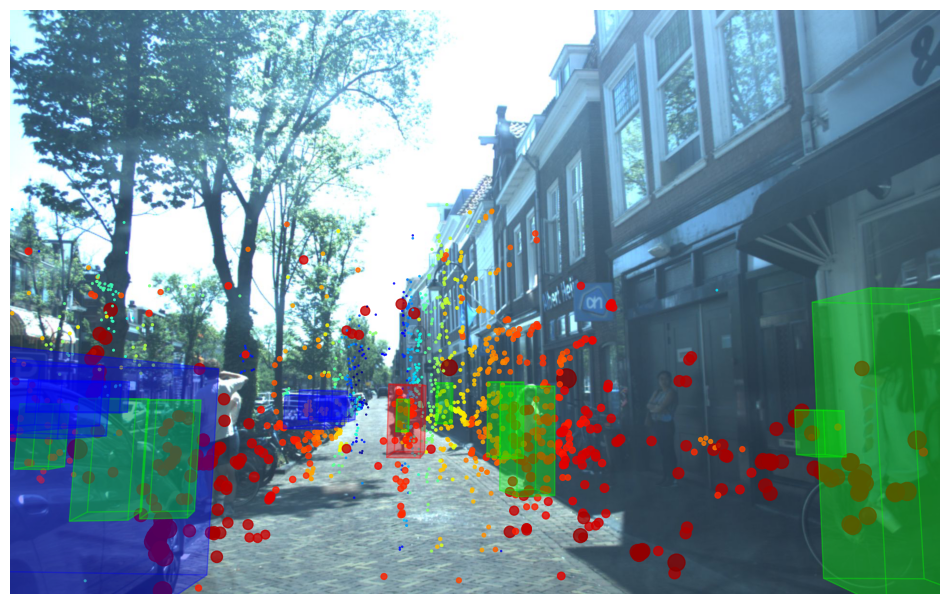}
		\caption{SECOND}
		\label{fig:second}
	\end{subfigure}
	\hfill
	\begin{subfigure}{0.375\textwidth}
		\centering 
		\includegraphics[width=\textwidth,trim={0 0 0 5.5cm},clip]{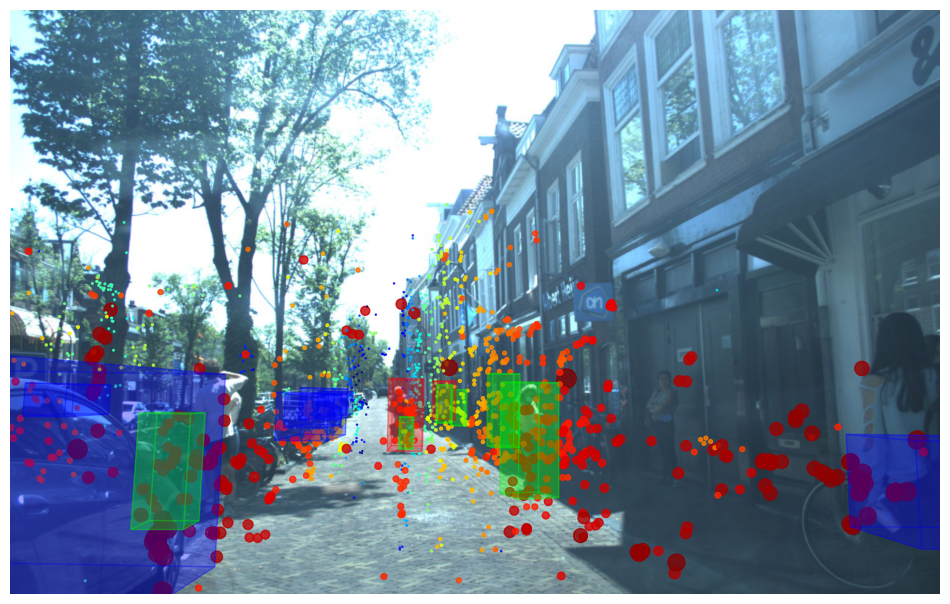}
		\caption{SECOND-MH}
		\label{fig:second-mh}
	\end{subfigure}
	\hfill
	\\
	\hfill
	\vspace{0.2cm}
	\begin{subfigure}{0.375\textwidth}
		\centering
		\includegraphics[width=\textwidth,trim={0 0 0 5.5cm},clip]{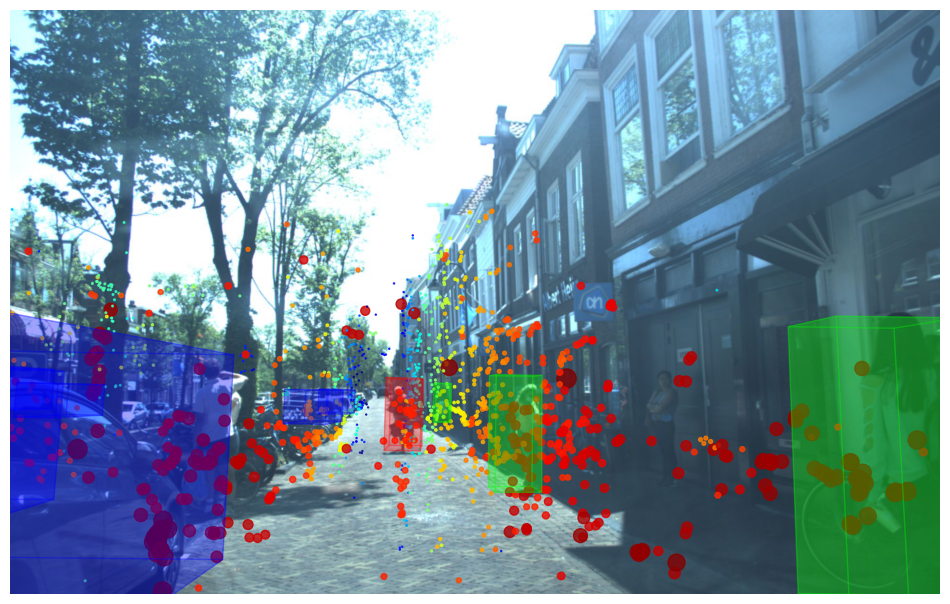}
		\caption{SECOND-IoU}
		\label{fig:second-iou}
	\end{subfigure}
	\hfill
	\begin{subfigure}{0.375\textwidth}
		\centering
		\includegraphics[width=\textwidth,trim={0 0 0 5.5cm},clip]{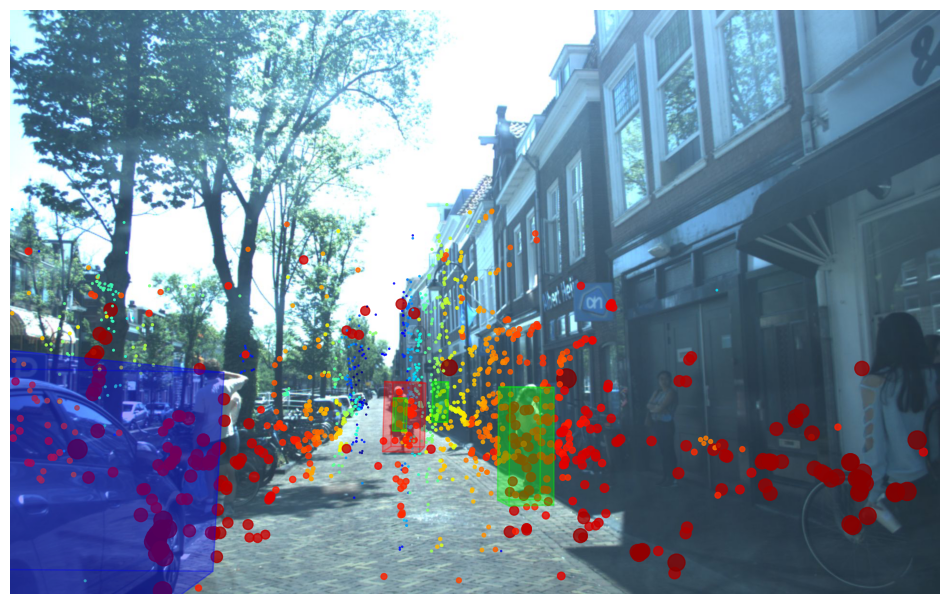}
		\caption{Part A\textsuperscript{2}}
		\label{fig:part_a2}
	\end{subfigure}
	\hfill
	\\
	\hfill
	\vspace{0.2cm}
	\begin{subfigure}{0.375\textwidth}
		\centering
		\includegraphics[width=\textwidth,trim={0 0 0 5.5cm},clip]{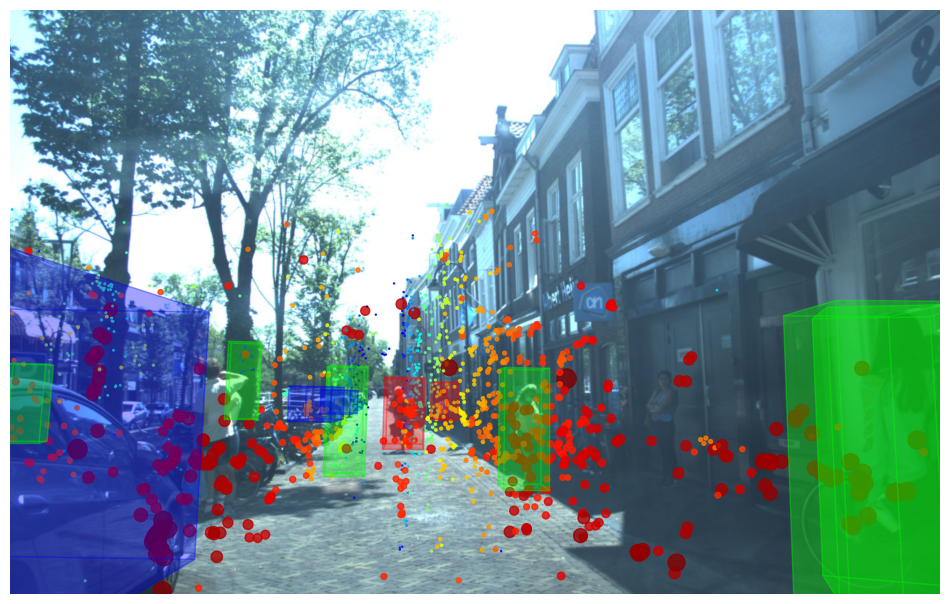}
		\caption{Voxel R-CNN}
		\label{fig:voxel-rcnn}
	\end{subfigure}	
	\hfill
	\begin{subfigure}{0.375\textwidth}
		\centering
		\includegraphics[width=\textwidth,trim={0 0 0 5.5cm},clip]{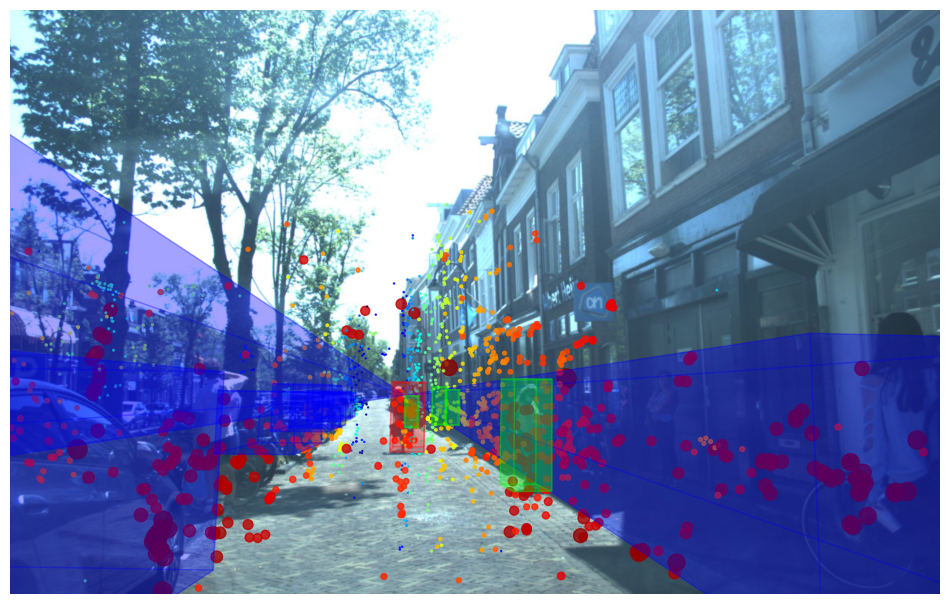}
		\caption{PointPillars-R}
		\label{fig:pointpillars-radar}
	\end{subfigure}
	\hfill
	\\
	\hfill
	\vspace{0.2cm}
	\begin{subfigure}{0.375\textwidth}
		\centering
		\includegraphics[width=\textwidth,trim={0 0 0 5.5cm},clip]{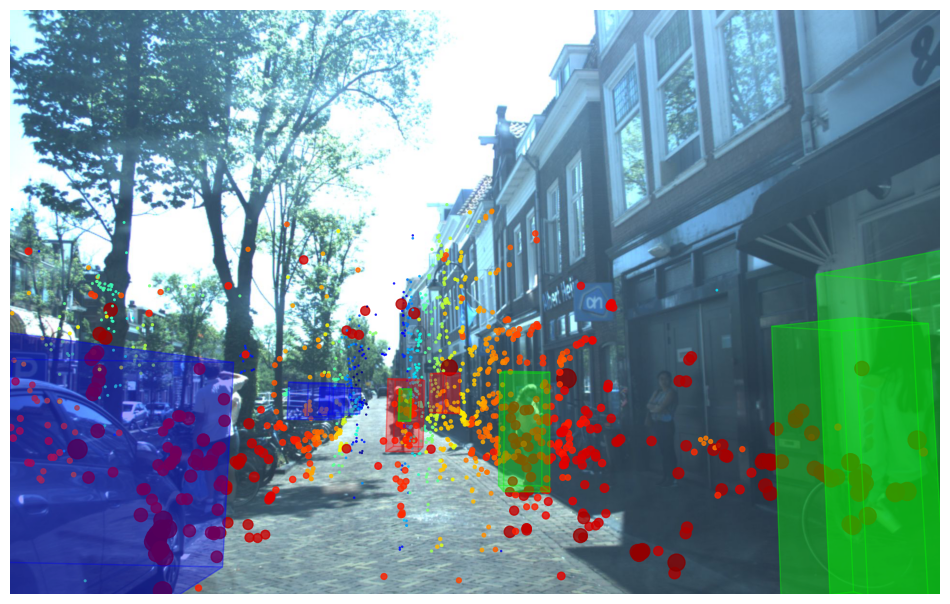}
		\caption{CenterPoint-R}
		\label{fig:CenterPoint}
	\end{subfigure}	
	\hfill
	\begin{subfigure}{0.375\textwidth}
		\centering
		\includegraphics[width=\textwidth,trim={0 0 0 5.5cm},clip]{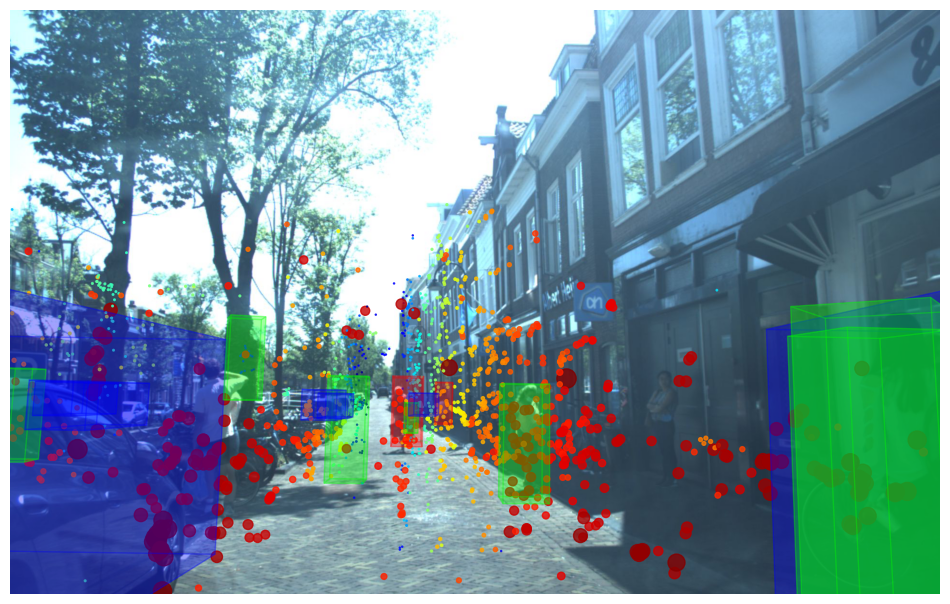}
		\caption{PV-RCNN}
		\label{fig:pv_rcnn_r}
	\end{subfigure}	
	\caption{Similar to \cref{fig:lidar_detections}, this visualization shows the camera image overlaid with the radar point cloud. The point color indicates the distance (red points represent close points, and blue ones are far away), whereas the size correlates with the RCS value (the larger the point, the larger the RCS). The shown radar points are accumulated over five frames, as \cite{ViewOfDelftDataset_22} identified this to be important to improve the detection results. As can be seen, the radar point cloud is much more sparse than the lidar point cloud, and from this snap shot data it is hard to visually distinguish clutter from detections from objects.}
	\label{fig:radar_detections}
\end{figure*}

\cref{fig:radar_detections} shows annotated ground truth and detection outputs on the radar data of the VoD data set from the models considered. As one would expect from the quantitative results in \cref{tab:comparison_radar_appendix} (and \cref{tab:comparison_radar} in the main paper), the detections are far from perfect and qualitatively inferior to the detection performance on the VoD lidar data, as can be seen in \cref{fig:lidar_detections}. Typical issues in \cref{fig:radar_detections} are \textit{correct} bounding boxes that are significantly off the true object, many false positive detections, and missed detections (false negatives). Detections that are far off could be improved by additionally estimating the ground plane and also considering the pitch and roll motion of the sensor vehicle that recorded the data. Due to extended passages of cobblestone streets in the inner city of Delft, this excitation is easily transmitted to the chassis affecting the sensor measurements, as can also be seen in the video data.

\begin{figure*}
	\centering
	\begin{subfigure}{0.49\textwidth}
		\centering
		\includegraphics[width=\textwidth]{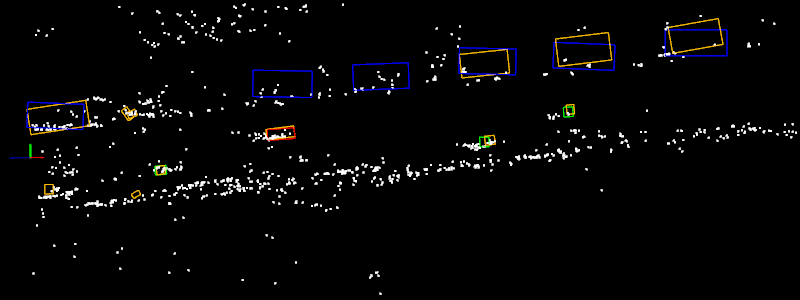}
		\caption{3DSSD}
		\label{fig:point-3dssd_BEV}
	\end{subfigure}
	\hfill
	\begin{subfigure}{0.49\textwidth}
		\centering
		\includegraphics[width=\textwidth]{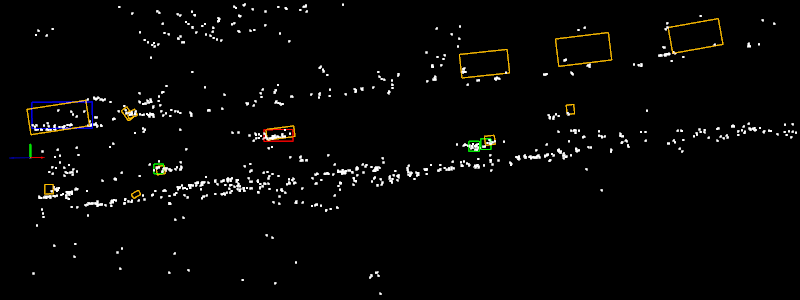}
		\caption{Point-RCNN}
		\label{fig:point-rcnn_BEV}
	\end{subfigure}
	\vskip\baselineskip
	\vspace{-0.2cm}
	\begin{subfigure}{0.49\textwidth}
		\centering 
		\includegraphics[width=\textwidth]{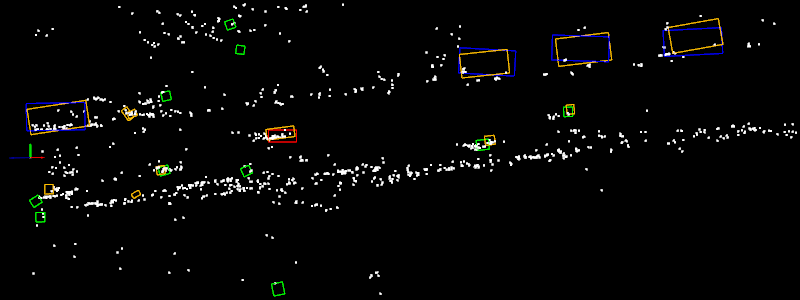}
		\caption{SECOND}
		\label{fig:second_BEV}
	\end{subfigure}
	\hfill
	\begin{subfigure}{0.49\textwidth}
		\centering 
		\includegraphics[width=\textwidth]{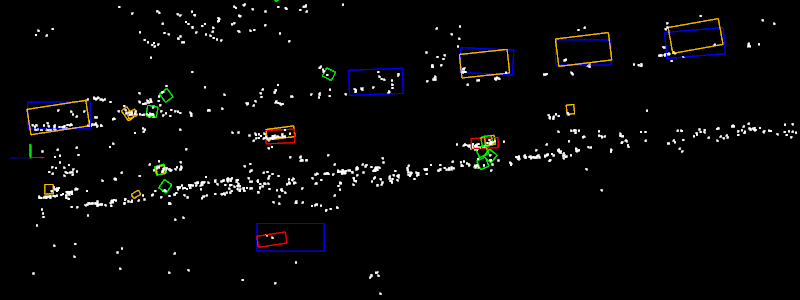}
		\caption{SECOND-MH}
		\label{fig:second-mh_BEV}
	\end{subfigure}
	\vskip\baselineskip
	\vspace{-0.2cm}
	\begin{subfigure}{0.49\textwidth}
		\centering
		\includegraphics[width=\textwidth]{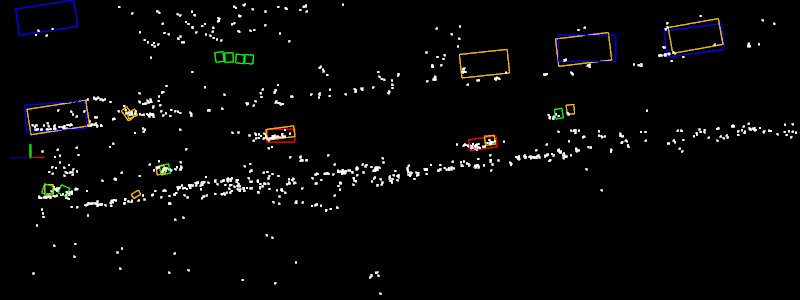}
		\caption{SECOND-IoU}
		\label{fig:second-iou_BEV}
	\end{subfigure}
	\hfill
	\begin{subfigure}{0.49\textwidth}
		\centering
		\includegraphics[width=\textwidth]{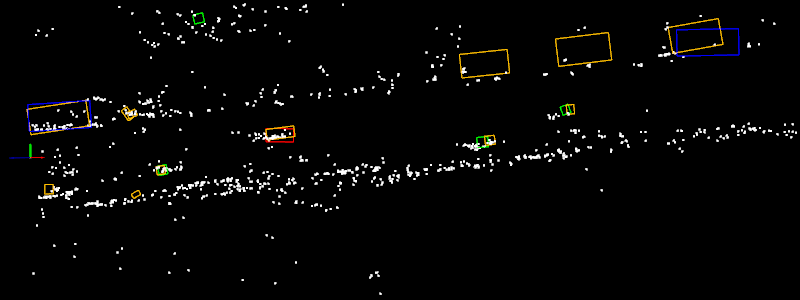}
		\caption{Part A\textsuperscript{2}}
		\label{fig:part_a2_BEV}
	\end{subfigure}
	\vskip\baselineskip
	\vspace{-0.2cm}
	\begin{subfigure}{0.49\textwidth}
		\centering
		\includegraphics[width=\textwidth]{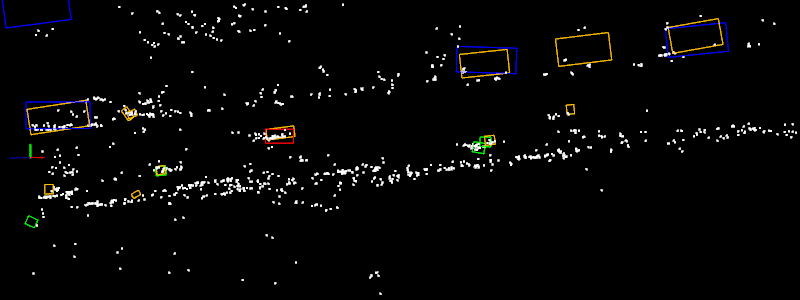}
		\caption{Voxel R-CNN}
		\label{fig:voxel-rcnn_BEV}
	\end{subfigure}	
	\hfill
	\begin{subfigure}{0.49\textwidth}
		\centering
		\includegraphics[width=\textwidth]{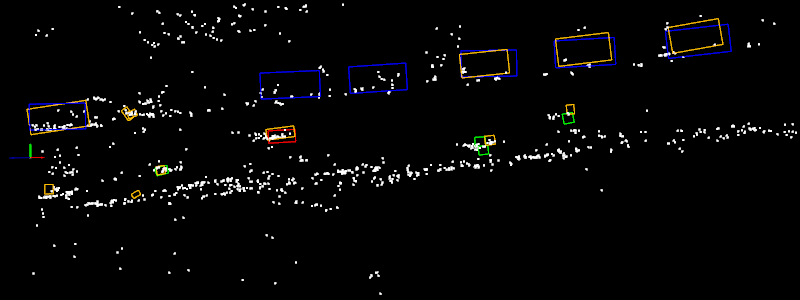}
		\caption{PointPillars-R}
		\label{fig:pointpillars-radar_BEV}
	\end{subfigure}
	\vskip\baselineskip
	\vspace{-0.2cm}
	\begin{subfigure}{0.49\textwidth}
		\centering
		\includegraphics[width=\textwidth]{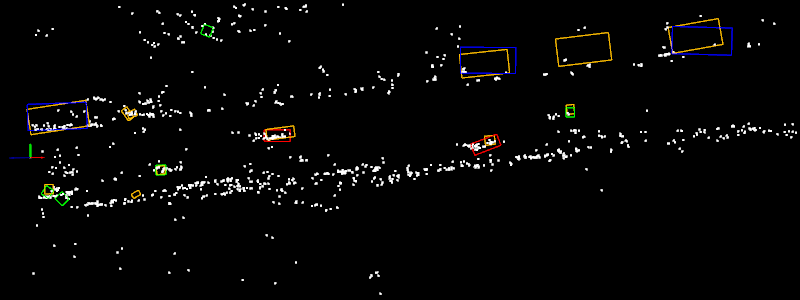}
		\caption{CenterPoint-R}
		\label{fig:CenterPoint_BEV}
	\end{subfigure}	
	\hfill
	\begin{subfigure}{0.49\textwidth}
		\centering
		\includegraphics[width=\textwidth]{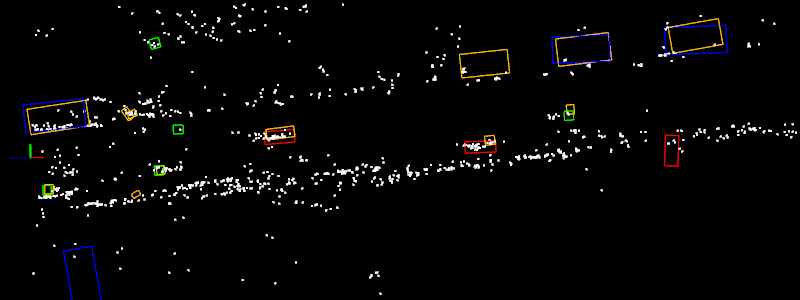}
		\caption{PV-RCNN}
		\label{fig:pv_rcnn_r_BEV}
	\end{subfigure}	
	\caption{The detection results from \cref{fig:radar_detections} are visualized in BEV-like representation. The white points are the single 3D radar measurements. Yellow cuboids represent the ground truth annotations for all object classes. Blue, red, and green cuboids visualize the detection outputs of the classes car, cyclist, and pedestrian, as in \cref{fig:radar_detections}. This view resolves possible overlays in \cref{fig:radar_detections} and emphasizes the sparsity of the radar point cloud.}
	\label{fig:radar_detections_BEV}
\end{figure*}

According to \cref{fig:radar_detections}, another frequent issue seems to be double-detections, \eg, correct cyclist detections overlaid by incorrect pedestrian detection. To resolve such object detections that seem to overlap with others, we additionally provide a BEV-like visualization in \cref{fig:radar_detections_BEV} containing radar points, ground truth annotations, and detections generated by the respective models. The additional visualization though shows that there are indeed almost no overlapping detections, as misleadingly indicated by the unfortunate perspective in \cref{fig:radar_detections}. Although, overlapping detections do not negatively affect the quantitative results since, for the calculation of the APs and the mAP only true positives (TPs) are considered.

\begin{figure*}
	\centering
	\begin{subfigure}{0.8\textwidth}
		\centering
		\includegraphics[width=\textwidth]{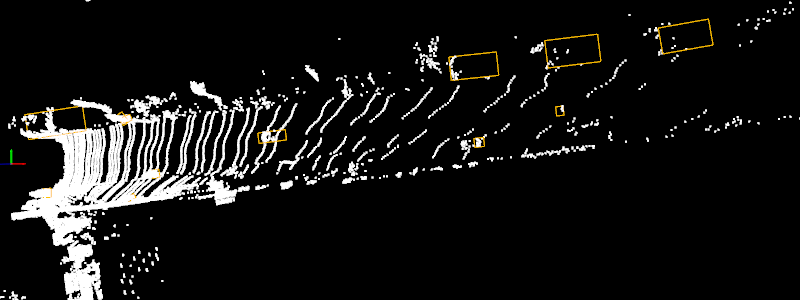}
		\caption{Lidar point cloud}
		\label{fig:lidar_BEV}
	\end{subfigure}
	\vskip\baselineskip
	\vspace{-0.2cm}
	\begin{subfigure}{0.8\textwidth}
		\centering 
		\includegraphics[width=\textwidth]{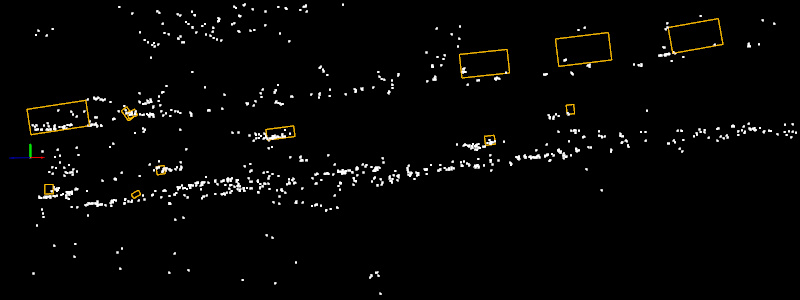}
		\caption{Radar point cloud accumulated over 5 frames}
		\label{fig:radar_5frames_BEV}
	\end{subfigure}
	\vskip\baselineskip
	\vspace{-0.2cm}
	\begin{subfigure}{0.8\textwidth}
		\centering
		\includegraphics[width=\textwidth]{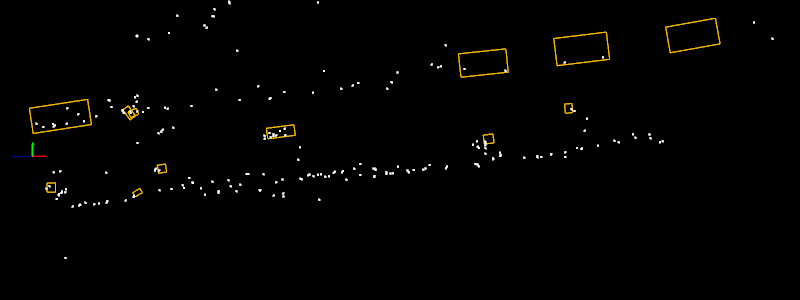}
		\caption{Radar point cloud of a single frame}
		\label{fig:radar_1frame_BEV}
	\end{subfigure}
	\caption{Single frame and accumulated point clouds from lidar and radar are visualized in a BEV-like representation. Yellow cuboids represent ground truth annotations for all object classes. The figures emphasize the sensors' characteristics concerning point cloud density (over distance), noise level of the measurements, and ability to capture different aspects of a traffic scene.}
	\label{fig:point_cloud_comparison_BEV}
\end{figure*}

As shown in \cref{fig:point_cloud_comparison_BEV}, even the accumulated radar point cloud is much sparser than the lidar point cloud. However, most of the points in the lidar point cloud are ground points. Ignoring those ground points, the accumulated radar point cloud is only much sparser at short distances when objects are close to the ego vehicle. At larger distances, the lidar point cloud is not much denser than the accumulated radar point cloud. However, the radar point clouds (accumulated and non-accumulated versions) are much noisier than the lidar point cloud. Therefore, both effects, the sparsity and the noise contribute to the degraded detection performance of radar based object detection compared to lidar based object detection.

\newpage

\begin{sidewaystable*}
	\centering
	\caption{Comparison of data sets (chronologically ordered) containing some form of radar data. Besides usual abbreviations like FoV (Field of View) or BB (Bounding Box), additional abbreviations are introduced to fit the table on a single page: Res. - Resolution; R - Range; A - Azimuth; E - Elevation; v - vertical; h - horizontal; PC - Point Cloud; Spec. - Spectra; Cam. - Camera; Dir. - Direction; Cls - Class; Anno. - Annotation; n.s. - not specified; y - yes; n - no; ff - front-facing.} 
	\begin{tabular}{@{\extracolsep{1pt}}p{2.625cm}p{0.875cm}p{0.975cm}p{1.875cm}p{1.75cm}p{0.5cm}p{0.5cm}p{0.5cm}p{0.25cm}p{0.5cm}p{0.8cm}p{0.375cm}p{0.875cm}p{0.125cm}p{0.125cm}p{0.375cm}p{0.625cm}}
		\toprule   
		{} & \multicolumn{2}{c}{Size} & \multicolumn{2}{c}{Radar Sensor Type} &\multicolumn{2}{c}{Observed Area} &  \multicolumn{2}{c}{Data Format} & \multicolumn{4}{c}{Additional Sensors} & \multicolumn{4}{c}{Annotations}\\
		\cmidrule{2-3} 
		\cmidrule{4-5} 
		\cmidrule{6-7} 
		\cmidrule{8-9} 
		\cmidrule{10-13}
		\cmidrule{14-17}
		Data Set & Length [\SI{}{\hour}]/[\SI{}{\kilo\meter}] & \#Scenes & Sensor & Res. R[\SI{}{\meter}]/A[\textdegree]/E[\textdegree] & View Range [\SI{}{\meter}] & FoV v[\textdegree]/h[\textdegree] & PC/ Spec. & Dop-pler & Cam. & Dir./ FoV [\textdegree] & Lidar & Dir./ FoV [\textdegree] & 2D BB & 3D BB & Point Cls Anno. & \#Cls \\ 
		\midrule
		nuScenes \cite{nuScenesDataset_20} & 5.55/n.s. & 1000 & Continental ARS-408-21 & n.s./n.s./n.s. & 250 & 360/- & y/n & n.s & y & -/360 & y & -/360 & y & y & y & 25 \\ 
		Lyft Level 5 \cite{LyftLevel5Dataset_20} & 1118.0/ 26000 & 170000 & n.s. & n.s./n.s./n.s. & - & 360/- & n.s. & n.s. & y & -/360 & y & -/360 & y & y & y & 3 n.s. \\
		Oxford Radar RobotCar \cite{OxfordRadarRobotCARDataset_19} & n.s./280 & n.s & Navtech CTS350-X & 0.04/0.9/n.s. & 163 & 360/- & n/y & n & y & -/360 & y & -/360 & n & n & n & - \\ 
		PREVENTION \cite{PREVENTIONDataset_19} & 6.0/540 & n.s. & Continental ARS-308 + SRR-208 & 2.0/1.0/n.s. 1.0/14.0/n.s. & 200, 50 & 56/-150/- & y/n & y & y & ff/ 48 & y & -/360 & y & y & y & 6 \\
		Zendar High Resolution Radar \cite{ZendarDataset_20} & n.s./n.s. & 27 & non-autom. grade sen. & 0.18/0.1/n.s. & 90 & 180/- & y/y & y & y & ff/ 60 & y & -/360 & y & n & n &  n.s. \\ 
		EU Long-term \cite{EULongTermDataset_20} & n.s./63.4 & 2 & Continental ARS-308 & 2.0/1.0/n.s. & 200 & 56/- & n.s. & y & y & ff/ 180 & y & -/360 & n.s. & n.s. & n.s. & n.s. \\
		RadarScenes \cite{RadarScenesDataset_21} & 4.3/100 & n.s. & series prod. autom. sen. & 0.15/0.5/n.s. & 100 & 270/- & y/n & y & y & ff/ 60 & - & - & n & n & n & 11+1 (5+1) \\
		RADIATE \cite{RADIATEDataset_21} & 3.0/n.s. & 7 & Navtech CTS350-X & 0.175/1.8/1.8 & 100 & 360/- & y/y & n & y & ff/ 60 & y & -/360 & y & n & y & 8 \\
		RaDICaL \cite{RaDICaLDataset_21} & n.s./n.s. & n.s. & Texas Instruments IWR1443 & 0.05-0.97/n.s./n.s. & 14.25-62.50 & 180/- & n/y & y & y & n.s. & - & - & y & n & n & n.s. \\
		Endeavour Radar \cite{EndeavourDataset_21} & 3.0/n.s. & n.s. & Continental ARS5430 & 0.39-1.79/1.6/n.s. & 70-250 & 360/- & y/n & y & - & -/- & y & corners/ 360 & n & n & y & n.s. \\
		RADIal \cite{RADIalDataset_22} & 2.0/n.s. & 91 & Valeo & 0.2/0.1/0.1 & 103 & 180/n.s. & y/y & y & y & ff/ 100 & y & ff/ 133 & y & n & y & 1 \\
		View-of-Delft \cite{ViewOfDelftDataset_22} & n.s./n.s. & n.s. & ZF FRGen 21 & 0.2/1.5/1.5 & 100 & 60/15 & y/n & y & y & ff/ 64 & y & -/360 & y & y & y & 13 \\
		\bottomrule
	\end{tabular}
	\label{tab:datasets}
\end{sidewaystable*}

\begin{table*}
	\centering
	\caption{Standard deviations of 3D object detection results on the VoD lidar data. For clarity reasons, we do not specify the mean in the supplementary material and refer to the main paper (\cref{tab:comparison_lidar}) for this information. Instead, we state the standard deviations here. This form of representing the results is repeated for \cref{tab:comparison_radar_appendix} (\cref{tab:comparison_radar} in the main paper) and \cref{tab:detailed_evaluation_second_appendix} (\cref{tab:detailed_evaluation_second} in the main paper), too}
	\begin{tabular}{@{\extracolsep{1pt}}p{2.25cm}p{0.475cm}p{0.475cm}p{0.475cm}p{0.55cm} | p{0.475cm}p{0.475cm}p{0.475cm}p{0.475cm}p{0.475cm}p{0.475cm}p{0.475cm}p{0.475cm}p{0.475cm}p{0.475cm}p{0.475cm}p{0.475cm}}
		\toprule   
		\multicolumn{1}{l}{Standard Dev.:} & \multicolumn{4}{c|}{mAP} & \multicolumn{4}{c}{Car} & \multicolumn{4}{c}{Pedestrian} & \multicolumn{4}{c}{Cyclist} \\
		\cmidrule{1-1}
		\cmidrule{2-5}
		\cmidrule{6-9}
		\cmidrule{10-13}
		\cmidrule{14-17}
		{} & \multicolumn{2}{c}{3D} & \multicolumn{2}{c|}{BEV} & \multicolumn{2}{c}{3D} & \multicolumn{2}{c}{BEV} & \multicolumn{2}{c}{3D} & \multicolumn{2}{c}{BEV} & \multicolumn{2}{c}{3D} & \multicolumn{2}{c}{BEV} \\
		\cmidrule{2-3}
		\cmidrule{4-5}
		\cmidrule{6-7}
		\cmidrule{8-9}
		\cmidrule{10-11}
		\cmidrule{12-13}
		\cmidrule{14-15}
		\cmidrule{16-17}
		Model & SR & MR & SR & MR & SR & MR & SR & MR & SR & MR & SR & MR & SR & MR & SR & MR \\ 
		\midrule
		3DSSD$^{\dagger}$ & $\pm$0.2 & $\pm$1.2 & $\pm$0.2 & $\pm$2.2 & $\pm$0.2 & $\pm$0.2 & $\pm$0.1 & $\pm$0.4 & $\pm$0.1 & $\pm$0.9 & $\pm$0.2 & $\pm$3.7 & $\pm$0.1 & $\pm$2.6 & $\pm$0.2 & $\pm$2.6 \\
		Point-RCNN & $\pm$1.6 & $\pm$1.6 & $\pm$1.7 & $\pm$2.7 & $\pm$0.2 & $\pm$0.3 & $\pm$0.2 & $\pm$3.7 & $\pm$0.2 & $\pm$4.0 & $\pm$0.4 & $\pm$3.9 & $\pm$4.4 & $\pm$0.6 & $\pm$4.4 & $\pm$0.6 \\
		\midrule
		SECOND$^{\dagger}$ & $\pm$2.1 & $\pm$0.7 & $\pm$1.4 & $\pm$1.3 & $\pm$0.1 & $\pm$0.3 & $\pm$3.6 & $\pm$0.3 & $\pm$2.9 & $\pm$0.3 & $\pm$0.6 & $\pm$1.8 & $\pm$3.2 & $\pm$1.4 & $\pm$0.1 & $\pm$1.7 \\
		SECOND-MH$^{\dagger}$ & $\pm$0.5 & $\pm$0.6 & $\pm$3.5 & $\pm$0.7 & $\pm$0.1 & $\pm$0.4 & $\pm$3.5 & $\pm$0.4 & $\pm$0.9 & $\pm$0.4 & $\pm$6.1 & $\pm$0.9 & $\pm$0.5 & $\pm$1.1 & $\pm$0.8 & $\pm$0.7 \\
		SECOND-IoU$^{\dagger}$ & $\pm$2.3 & $\pm$2.7 & $\pm$2.6 & $\pm$1.2 & $\pm$0.1 & $\pm$2.2 & $\pm$3.6 & $\pm$0.4 & $\pm$2.9 & $\pm$2.8 & $\pm$0.6 & $\pm$1.5 & $\pm$3.8 & $\pm$3.0 & $\pm$1.6 & $\pm$1.7 \\
		Part A\textsuperscript{2} & $\pm$1.7 & $\pm$1.7 & $\pm$1.3 & $\pm$1.5 & $\pm$0.2 & $\pm$2.0 & $\pm$2.9 & $\pm$1.0 & $\pm$1.0 & $\pm$1.1 & $\pm$0.5 & $\pm$1.8 & $\pm$3.9 & $\pm$1.9 & $\pm$0.5 & $\pm$1.6 \\
		Voxel R-CNN & $\pm$0.3 & $\pm$1.4 & $\pm$0.1 & $\pm$0.9 & $\pm$0.2 & $\pm$0.3 & $\pm$0.0 & $\pm$0.2 & $\pm$0.4 & $\pm$3.1 & $\pm$0.0 & $\pm$1.1 & $\pm$0.3 & $\pm$0.9 & $\pm$0.3 & $\pm$1.4 \\
		\midrule
		PointPillars-L$^{\dagger}$ & $\pm$0.5 & $\pm$1.5 & $\pm$1.4 & $\pm$1.9 & $\pm$0.3 & $\pm$0.3 & $\pm$3.0 & $\pm$0.4 & $\pm$0.9 & $\pm$0.8 & $\pm$0.7 & $\pm$2.3 & $\pm$0.5 & $\pm$3.3 & $\pm$0.5 & $\pm$3.1 \\
		CenterPoint-L & $\pm$1.5 & $\pm$1.1 & $\pm$1.7 & $\pm$1.1 & $\pm$1.3 & $\pm$1.2 & $\pm$0.4 & $\pm$0.6 & $\pm$3.0 & $\pm$1.7 & $\pm$1.6 & $\pm$1.6 & $\pm$0.2 & $\pm$0.4 & $\pm$3.2 & $\pm$1.1 \\
		\midrule
		PV-RCNN & $\pm$1.9 & $\pm$3.0 & $\pm$0.1 & $\pm$0.9 & $\pm$4.3 & $\pm$3.3 & $\pm$4.3 & $\pm$0.4 & $\pm$0.3 & $\pm$2.1 & $\pm$2.2 & $\pm$2.2 & $\pm$2.5 & $\pm$2.9 & $\pm$2.5 & $\pm$3.0 \\
		\bottomrule
	\end{tabular}
	\label{tab:comparison_lidar_appendix}
\end{table*}

\begin{table*}
	\centering
	\caption{Standard deviations of 3D object detection results on the VoD radar data. The numbers represent the standard deviation. The mean values are stated in the main paper in \cref{tab:comparison_radar}. As done in \cref{tab:detailed_evaluation_second_appendix} too, we rounded double-digit standard deviation values to integer numbers to fit them in table well.}
	\begin{tabular}{@{\extracolsep{1pt}}p{2.25cm}p{0.475cm}p{0.475cm}p{0.475cm}p{0.55cm} | p{0.475cm}p{0.475cm}p{0.475cm}p{0.475cm}p{0.475cm}p{0.475cm}p{0.475cm}p{0.475cm}p{0.475cm}p{0.475cm}p{0.475cm}p{0.475cm}}
		\toprule   
		\multicolumn{1}{l}{Standard Dev.:} & \multicolumn{4}{c|}{mAP} & \multicolumn{4}{c}{Car} & \multicolumn{4}{c}{Pedestrian} & \multicolumn{4}{c}{Cyclist} \\
		\cmidrule{1-1}
		\cmidrule{2-5}
		\cmidrule{6-9}
		\cmidrule{10-13}
		\cmidrule{14-17}
		{} & \multicolumn{2}{c}{3D} & \multicolumn{2}{c|}{BEV} & \multicolumn{2}{c}{3D} & \multicolumn{2}{c}{BEV} & \multicolumn{2}{c}{3D} & \multicolumn{2}{c}{BEV} & \multicolumn{2}{c}{3D} & \multicolumn{2}{c}{BEV} \\
		\cmidrule{2-3}
		\cmidrule{4-5}
		\cmidrule{6-7}
		\cmidrule{8-9}
		\cmidrule{10-11}
		\cmidrule{12-13}
		\cmidrule{14-15}
		\cmidrule{16-17}
		Model & SR & MR & SR & MR & SR & MR & SR & MR & SR & MR & SR & MR & SR & MR & SR & MR \\ 
		\midrule
		3DSSD$^{\dagger}$ & $\pm$3.7 & $\pm$3.6 & $\pm$3.4 & $\pm$3.9 & $\pm$4.5 & $\pm$4.2 & $\pm$4.4 & $\pm$5.0 & $\pm$1.5 & $\pm$4.0 & $\pm$0.7 & $\pm$2.7 & $\pm$4.9 & $\pm$2.5 & $\pm$5.0 & $\pm$4.1 \\
		Point-RCNN & $\pm$3.7 & $\pm$1.8 & $\pm$2.2 & $\pm$2.7 & $\pm$4.8 & $\pm$0.0 & $\pm$4.6 & $\pm$3.6 & $\pm$1.3 & $\pm$2.6 & $\pm$1.0 & $\pm$4.1 & $\pm$5.1 & $\pm$2.6 & $\pm$1.0 & $\pm$0.3 \\
		\midrule
		SECOND$^{\dagger}$ & $\pm$2.8 & $\pm$3.1 & $\pm$1.5 & $\pm$1.9 & $\pm$2.9 & $\pm$0.5 & $\pm$0.9 & $\pm$0.9 & $\pm$1.0 & $\pm$4.3 & $\pm$1.2 & $\pm$4.1 & $\pm$4.5 & $\pm$4.5 & $\pm$2.6 & $\pm$0.7 \\
		SECOND-MH$^{\dagger}$ & $\pm$1.5 & $\pm$2.3 & $\pm$1.6 & $\pm$2.6 & $\pm$0.8 & $\pm$2.6 & $\pm$1.1 & $\pm$4.0 & $\pm$0.8 & $\pm$3.9 & $\pm$0.4 & $\pm$2.6 & $\pm$2.9 & $\pm$0.4 & $\pm$3.3 & $\pm$1.4 \\
		SECOND-IoU$^{\dagger}$ & $\pm$2.4 & $\pm$1.2 & $\pm$1.7 & $\pm$4.1 & $\pm$2.9 & $\pm$2.2 & $\pm$1.5 & $\pm$4.7 & $\pm$0.9 & $\pm$0.3 & $\pm$0.4 & $\pm$1.2 & $\pm$3.3 & $\pm$1.2 & $\pm$3.3 & $\pm$6.5 \\
		Part A\textsuperscript{2} & $\pm$0.9 & $\pm$2.8 & $\pm$1.8 & $\pm$2.7 & $\pm$0.5 & $\pm$0.9 & $\pm$0.1 & $\pm$0.6 & $\pm$0.1 & $\pm$3.6 & $\pm$0.5 & $\pm$2.7 & $\pm$2.2 & $\pm$4.0 & $\pm$4.7 & $\pm$4.9 \\
		Voxel R-CNN & $\pm$0.4 & $\pm$1.9 & $\pm$1.5 & $\pm$2.6 & $\pm$0.2 & $\pm$1.0 & $\pm$1.0 & $\pm$1.6 & $\pm$0.4 & $\pm$3.9 & $\pm$1.7 & $\pm$4.6 & $\pm$0.6 & $\pm$0.7 & $\pm$1.8 & $\pm$1.5 \\
		\midrule
		PointPillars-R$^{\dagger}$ & $\pm$2.7 & $\pm$3.5 & $\pm$0.9 & $\pm$2.7 & $\pm$3.2 & $\pm$0.1 & $\pm$0.6 & $\pm$2.7 & $\pm$0.3 & $\pm$0.5 & $\pm$1.4 & $\pm$0.7 & $\pm$4.5 & $\pm$9.3 & $\pm$0.8 & $\pm$4.7 \\
		CenterPoint-R & $\pm$1.3 & $\pm$1.7 & $\pm$3.8 & $\pm$4.3 & $\pm$0.6 & $\pm$2.2 & $\pm$4.5 & $\pm$5.9 & $\pm$1.9 & $\pm$0.4 & $\pm$1.4 & $\pm$3.4 & $\pm$1.3 & $\pm$2.6 & $\pm$5.5 & $\pm$3.5 \\
		\midrule
		PV-RCNN & $\pm$1.0 & $\pm$1.3 & $\pm$4.7 & $\pm$13 & $\pm$1.7 & $\pm$2.4 & $\pm$2.0 & $\pm$10 & $\pm$0.2 & $\pm$0.3 & $\pm$3.3 & $\pm$12 & $\pm$1.2 & $\pm$1.2 & $\pm$8.8 & $\pm$17 \\
		\bottomrule
	\end{tabular}
	\label{tab:comparison_radar_appendix}
\end{table*}

\begin{table*}
	\centering
	\caption{Object detection results on the Astyx data for the car class only. Since other object classes have a low occurrence rate, only this class is evaluated.} 
	\begin{tabular}{@{\extracolsep{5pt}}p{2.75cm}p{1.5cm}p{1.5cm}p{1.5cm}p{1.5cm}p{1.5cm}p{1.5cm}}
		\toprule   
		{} & \multicolumn{3}{c}{3D} & \multicolumn{3}{c}{BEV} \\
		\cmidrule{2-4}
		\cmidrule{5-7}
		Model & SR & MR & LR & SR & MR & LR \\ 
		\midrule
		3DSSD$^{\dagger}$ & 17.5$\pm$0.5 & 4.6$\pm$2.4 & 3.6$\pm$1.6 & 34.1$\pm$0.4 & 14.7$\pm$1.3 & 7.1$\pm$4.0 \\
		Point-RCNN & 2.5$\pm$1.5 & 0.4$\pm$0.3 & 0.2$\pm$0.3 & 8.7$\pm$3.7 & 3.0$\pm$0.5 & 0.3$\pm$0.3 \\
		\midrule
		SECOND$^{\dagger}$ & 13.0$\pm$7.0 & 6.1$\pm$5.5 & 1.1$\pm$0.9 & 25.0$\pm$9.7 & 19.4$\pm$11.8 & 12.3$\pm$5.7 \\
		SECOND-MH$^{\dagger}$ & 19.7$\pm$6.8 & 9.6$\pm$4.5 & 2.6$\pm$1.8 & 40.2$\pm$7.3 & 24.6$\pm$10.5 & 15.2$\pm$8.6 \\
		SECOND-IoU$^{\dagger}$ & 20.1$\pm$1.6 & 8.3$\pm$1.9 & 4.2$\pm$1.9 & 35.1$\pm$0.8 & 25.4$\pm$2.3  & \textbf{16.4}$\pm$3.4 \\
		Part A\textsuperscript{2} & 9.9$\pm$4.0  & 2.1$\pm$1.0 & 1.0$\pm$0.5 & 19.9$\pm$6.0 & 7.5 $\pm$0.8  & 6.0 $\pm$0.8 \\
		Voxel R-CNN & 20.9$\pm$4.8 & 6.4$\pm$1.5 &  1.4$\pm$0.8 &  38.7$\pm$8.7 &  21.0$\pm$7.8 &  11.4$\pm$1.6 \\
		\midrule
		PointPillars-R$^{\dagger}$ & 14.1$\pm$1.5 & 2.2$\pm$2.8 & 0.2$\pm$0.3 & \textbf{40.8}$\pm$4.9 & 22.2$\pm$1.4 & 13.9$\pm$3.4\\
		CenterPoint-R & 22.6$\pm$4.0 & \textbf{10.4}$\pm$1.8 & \textbf{6.1}$\pm$2.6 & 37.6$\pm$2.4 & 21.9$\pm$4.1 & 9.2$\pm$1.2 \\
		\midrule
		PV-RCNN & \textbf{24.4}$\pm$2.8 & 9.0$\pm$1.9 & 3.0$\pm$1.3 & 39.8$\pm$1.0 & \textbf{26.7}$\pm$3.5 & 15.4$\pm$2.8 \\
		\bottomrule
	\end{tabular}
	\label{tab:astyx_evaluation_appendix}
\end{table*}

\newpage

\begin{table*}
	\centering
	\caption{Standard deviations of 3D object detection results on the Astyx data. The mean values are stated in the main paper in \cref{tab:detailed_evaluation_second}. We investigate small voxel and pillar sizes (svs, sps) vs. large voxel and pillar sizes (lvs, lps). For the adapted SECOND model with lvs we additionally experimented with an adapted learning rate scheduler due to the insights from the training of the initial model.} 
	\begin{tabular}{@{\extracolsep{1pt}}p{2.125cm}p{0.475cm}p{0.475cm}p{0.475cm}p{0.55cm} | p{0.475cm}p{0.475cm}p{0.475cm}p{0.475cm}p{0.475cm}p{0.475cm}p{0.475cm}p{0.475cm}p{0.475cm}p{0.475cm}p{0.475cm}p{0.475cm}}
		\toprule   
		\multicolumn{1}{l}{Standard Dev.:} & \multicolumn{4}{c|}{mAP} & \multicolumn{4}{c}{Car} & \multicolumn{4}{c}{Pedestrian} & \multicolumn{4}{c}{Cyclist} \\
		\cmidrule{1-1}
		\cmidrule{2-5}
		\cmidrule{6-9}
		\cmidrule{10-13}
		\cmidrule{14-17}
		{} & \multicolumn{2}{c}{3D} & \multicolumn{2}{c|}{BEV} & \multicolumn{2}{c}{3D} & \multicolumn{2}{c}{BEV} & \multicolumn{2}{c}{3D} & \multicolumn{2}{c}{BEV} & \multicolumn{2}{c}{3D} & \multicolumn{2}{c}{BEV} \\
		\cmidrule{2-3}
		\cmidrule{4-5}
		\cmidrule{6-7}
		\cmidrule{8-9}
		\cmidrule{10-11}
		\cmidrule{12-13}
		\cmidrule{14-15}
		\cmidrule{16-17}
		Model & SR & MR & SR & MR & SR & MR & SR & MR & SR & MR & SR & MR & SR & MR & SR & MR \\ 
		\midrule
		PointPillars-R & $\pm$2.7 & $\pm$3.5 & $\pm$0.9 & $\pm$2.7 & $\pm$3.2 & $\pm$0.1 & $\pm$0.6 & $\pm$2.7 & $\pm$0.3 & $\pm$0.5 & $\pm$1.4 & $\pm$0.7 & $\pm$12 & $\pm$9.3 & $\pm$0.8 & $\pm$4.7 \\	
		\midrule
		PointPillars-R (sps) & $\pm$2.7 & $\pm$3.1 & $\pm$1.9 & $\pm$2.8 & $\pm$2.8 & $\pm$3.0 & $\pm$1.2 & $\pm$2.9 & $\pm$1.6 & $\pm$4.7 & $\pm$1.5 & $\pm$2.4 & $\pm$3.5 & $\pm$1.6 & $\pm$3.1 & $\pm$3.0 \\
		\midrule
		\midrule
		SECOND & $\pm$2.8 & $\pm$3.1 & $\pm$1.5 & $\pm$1.9 & $\pm$2.9 & $\pm$0.5 & $\pm$0.9 & $\pm$0.9 & $\pm$1.0 & $\pm$4.3 & $\pm$1.2 & $\pm$4.1 & $\pm$4.5 & $\pm$4.5 & $\pm$2.6 & $\pm$0.7 \\
		\midrule
		SECOND (lvs) & $\pm$3.6 & $\pm$3.2 & $\pm$4.9 & $\pm$6.3 & $\pm$0.7 & $\pm$1.7 & $\pm$2.2 & $\pm$4.1 & $\pm$1.7 & $\pm$4.2 & $\pm$3.6 & $\pm$3.5 & $\pm$8.5 & $\pm$3.7 & $\pm$9.0 & $\pm$11 \\
		SECOND (lvs, lr scheduler) & $\pm$1.5 & $\pm$3.1 & $\pm$0.8 & $\pm$2.7 & $\pm$0.4 & $\pm$0.7 & $\pm$0.2 & $\pm$1.0 & $\pm$0.6 & $\pm$3.5 & $\pm$2.1 & $\pm$3.9 & $\pm$3.4 & $\pm$5.0 & $\pm$0.3 & $\pm$3.1 \\
		\bottomrule
	\end{tabular}
	\label{tab:detailed_evaluation_second_appendix}
\end{table*}

\newpage

% from: https://www.reddit.com/r/LaTeX/comments/50edjz/subfigure_odd_number_of_plots/

\newpage

\newpage

\newpage

\clearpage

\end{document}